\documentclass{article}
\usepackage[numbers,sort&compress]{natbib}
\usepackage{neurips_2024}
\usepackage{amsmath, amsfonts, amssymb}
\usepackage{booktabs}
\usepackage{graphicx}
\usepackage{wrapfig}
\usepackage{tikz}
\usetikzlibrary{shapes.geometric, arrows.meta, positioning, calc, backgrounds, fit}
\usepackage{hyphenat}
\usepackage[utf8]{inputenc}
\usepackage{microtype}
\usepackage{xcolor}
\usepackage{subcaption}

\title{Motif-Video 2B: Technical Report}
\author{\textbf{Motif Technologies}\thanks{This work was conducted independently on Microsoft Azure, using compute resources separate from those supported by the Korea Sovereign AI Foundation Model (K-AI) project. Infrastructure was managed with SkyPilot~\citep{yang2023skypilot} on a Kubernetes cluster running on Azure nodes.}}

\begin{document}
\maketitle

\begin{abstract}

Training strong video generation models usually requires massive datasets, large parameter counts, and substantial compute. In this work, we ask whether strong text-to-video quality is possible at a much smaller budget: fewer than 10M clips and less than 100,000 H200 GPU hours.
Our core claim is that part of the answer lies in how model capacity is organized, not only in how much of it is used. In video generation, prompt alignment, temporal consistency, and fine-detail recovery can interfere with one another when they are handled through the same pathway. Motif-Video 2B addresses this by separating these roles architecturally, rather than relying on scale alone.
The model combines two key ideas. First, Shared Cross-Attention strengthens text control when video token sequences become long. Second, a three-part backbone separates early fusion, joint representation learning, and detail refinement. To make this design effective under a limited compute budget, we pair it with an efficient training recipe based on dynamic token routing and early-phase feature alignment to a frozen pretrained video encoder.
Our analysis shows that later blocks develop clearer cross-frame attention structure than standard single-stream baselines. On VBench, Motif-Video~2B reaches 83.76\%, surpassing Wan2.1 14B while using 7$\times$ fewer parameters and substantially less training data. These results suggest that careful architectural specialization, combined with an efficiency-oriented training recipe, can narrow or exceed the quality gap typically associated with much larger video models.

\end{abstract}


\section{Introduction}
\label{sec:introduction}

Video generation has entered a scaling regime. The most capable open models, Wan2.1~\citep{wan2025wan}, HunyuanVideo~\citep{kong2024hunyuanvideo}, and Seedance~\citep{gao2025seedance}, are trained on hundreds of millions of curated clips, with parameter counts ranging from 5B to 14B. This concentration of resources has produced impressive results, but it has also narrowed participation: in practice, training a competitive video generation model is accessible to very few groups.

The image generation domain has begun to challenge this assumption. Earlier PixArt-$\alpha$~\citep{chen2023pixart} efforts and the later PRX-3 project~\citep{photoroom2025prx}, together with related ImageNet speedrun efforts~\citep{bhanded2025speedrunning}, show that careful engineering can partially substitute for brute-force scale. In particular, representation alignment, token routing, and principled architectural choices can produce competitive image generation models within a single day of training on modest hardware. The natural question is whether this philosophy transfers to video.

Video is harder than image generation because the model must satisfy three goals at once: (1) follow the text prompt, (2) keep motion and content consistent across frames, and (3) recover fine visual details. We refer to the resulting competition for shared model capacity as \textit{objective interference}. In practice, improvements along one dimension can come at the expense of another.

As sequence length increases, text tokens become sparse relative to video tokens, which weakens text control in standard cross-attention. At the same time, learning long-range temporal structure can conflict with per-frame detail synthesis. A frozen visual encoder can help in early training, but later it can limit adaptation to the target distribution. As a result, scaling model size and data often delays these tensions rather than resolving them directly. Our central hypothesis is that objective interference is better addressed by explicit role separation than by scaling alone. We test this as an exploratory design hypothesis, not a strict causal claim, and use per-component attention analysis as supporting evidence.

Following this hypothesis, we build \textbf{Motif-Video 2B}, a text-to-video diffusion transformer. More broadly, the paper asks whether architectural specialization, combined with an efficient training recipe, can substitute in part for brute-force scale in video generation. The overall architecture follows a three-stage layout: dual-stream blocks for initial text-video fusion, single-stream blocks for joint representation learning, and DDT~\citep{wang2025ddt} blocks for decoupled semantic encoding and detail decoding. This extends the functional role-separation philosophy of FLUX~\citep{flux2024} into the spatiotemporal domain.

\begin{figure*}[t]
    \centering
    \includegraphics[width=\textwidth]{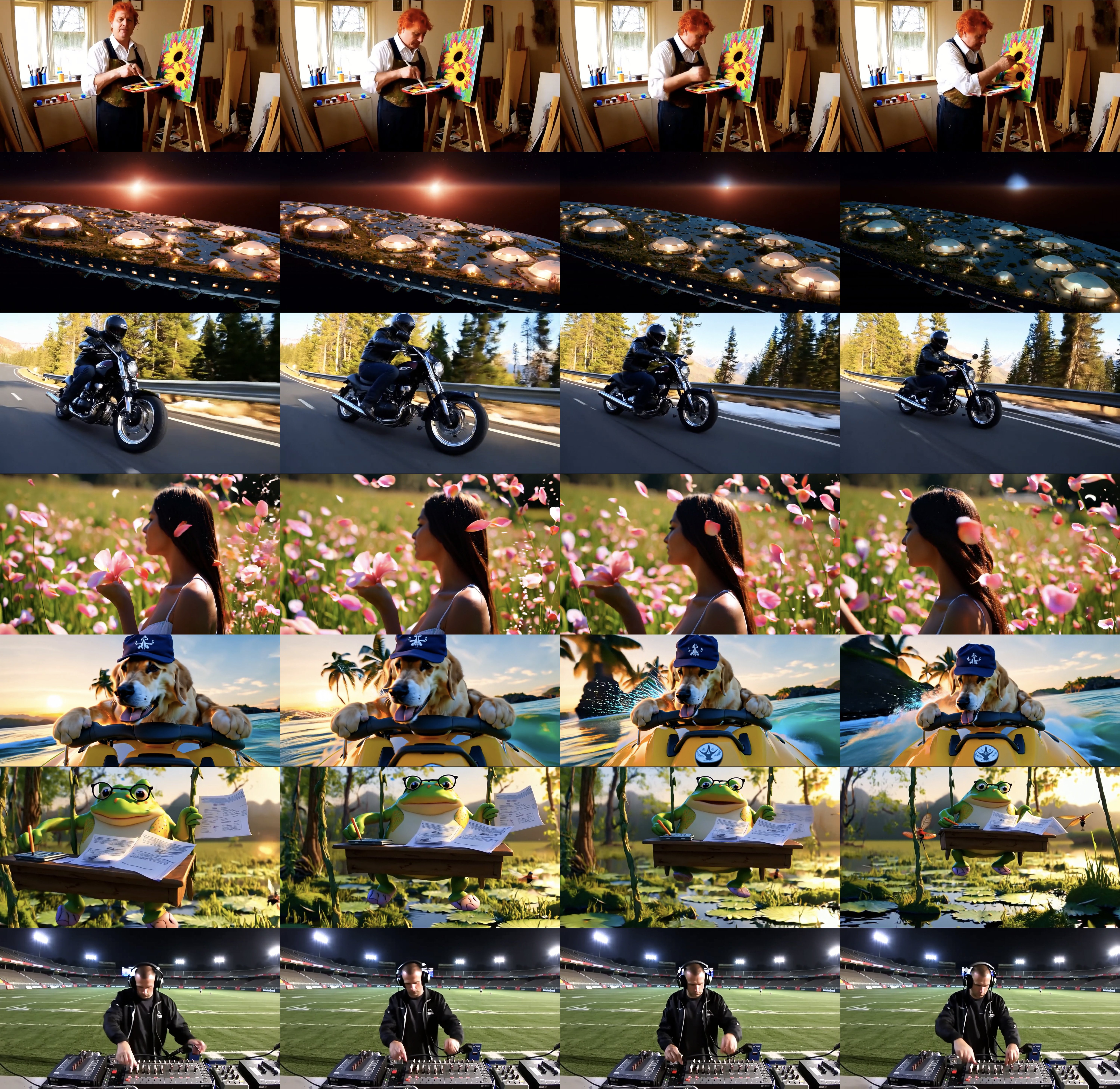}
    \caption{\textbf{Representative generations from Motif-Video 2B.} Frames are captured from videos generated by our 2B-parameter text-to-video model across a diverse set of prompts, illustrating the combination of prompt fidelity, temporal coherence, and visual detail that we target throughout this work. The banner is intended as a qualitative teaser; later sections analyze the architectural and training choices that make these generations possible under a micro-budget training regime.}
    \label{fig:main_banner}
\end{figure*}

Within the single-stream stage, we observe that standard self-attention insufficiently preserves text alignment as sequence length grows: text tokens become relatively sparse in the joint attention matrix, and their influence degrades. We address this with \textbf{Shared Cross-Attention}, which constructs cross-attention keys and values by reusing weights already learned by the self-attention pathway, constraining text-video attention to operate within the model's existing representation manifold. On the training side, we compose a micro-budget recipe that combines TREAD token routing~\citep{krause2025tread} and early-phase REPA with a V-JEPA teacher~\citep{yu2024representation,bardes2024revisiting}. To our knowledge, this combination has not previously been applied to text-to-video training. Data quality is controlled through a learned preference model over our 2.8M-clip proprietary collection.

Per-component attention pattern analysis supports our design intent: DDT blocks exhibit clear inter-frame attention structure that is absent in single-stream layers, and Shared Cross-Attention shows measurably more stable text-region activation across long sequences. Trained on fewer than 10M clips within 100,000 GPU hours, Motif-Video 2B achieves 83.76\% on the VBench leaderboard~\citep{huang2024vbench}, surpassing Wan2.1-14B at 7$\times$ fewer parameters and an order of magnitude less training data.

We summarize our contributions as follows:

\begin{itemize}

\item We present Shared Cross-Attention, a residual cross-attention mechanism that shares self-attention K--V weights to stabilize text--video alignment under long-context token sparsity, and show that it measurably corrects the alignment degradation observed in standard cross-attention at extended sequence lengths.
\item We introduce DDT and TREAD to video generation, and show through attention pattern analysis that the condition encoder develops inter-frame attention structure in the video setting, an inductive bias for temporal coherence that motivates the three-stage architectural layout.
\item We demonstrate that a micro-budget training recipe, combining TREAD token routing and early-phase REPA with a V-JEPA teacher, is sufficient to train a 2B model on fewer than 10M clips that reaches 83.76\% on VBench, surpassing Wan2.1-14B.

\end{itemize}


\section{Related Work}
\label{sec:related_work}

\paragraph{Production-scale video generation.}
The current landscape of text-to-video generation is defined by models trained at substantial scale. Open models such as CogVideoX~\citep{yang2024cogvideox}, Wan2.1~\citep{wan2025wan}, Wan2.2, HunyuanVideo~\citep{kong2024hunyuanvideo}, HunyuanVideo 1.5~\citep{wu2025hunyuanvideo}, Waver~\citep{zhang2025waver}, and Seedance~\citep{gao2025seedance, seedance2025seedance} are trained on data pools of hundreds of millions of video clips, with parameter counts ranging from 5B to 14B. Proprietary systems including Sora, Veo 3, Kling, Runway Gen-4, and Grok Aurora appear to operate at comparable or larger scale, although their training details are largely undisclosed. Despite substantial architectural diversity across these systems, their reported performance has largely been achieved in a regime of very large data and model scale. This work asks whether competitive quality can also be reached under a much smaller training budget.

\paragraph{Video and image diffusion transformer architectures.}
The MMDiT design of SD3 and FLUX established the dual-stream / single-stream split as a principled approach to modality-aware processing: early layers maintain separate text and image streams to avoid premature feature entanglement, while later layers merge them for joint generation~\citep{flux2024, esser2024scaling}. CogVideoX extends this idea to video through joint 3D attention over text and video tokens, while Seedance revisits stream separation from a different perspective. DDT, originally proposed for image generation, addresses the tension between low-frequency semantic encoding and high-frequency detail decoding by decoupling these roles into an explicit encoder-decoder design. These works suggest that architectural role separation can be a useful inductive bias, but they do not directly address the long-context text-alignment problem that becomes pronounced in text-to-video generation as frame count increases.

\paragraph{Efficient training for diffusion models.}
Significant progress has been made on reducing the cost of diffusion model training in the image domain. REPA aligns early DiT hidden states with a frozen visual encoder and substantially accelerates convergence on ImageNet; follow-up work shows that this benefit is concentrated in early training and advocates disabling the alignment objective later to avoid a capacity bottleneck~\citep{yu2024representation}. TREAD routes a subset of tokens from shallow to deep layers during training, reducing FLOPs while providing early layers with deeper supervision~\citep{krause2025tread}. The PRX-3 project and related ImageNet speedrun efforts combine such ideas into micro-budget training recipes that achieve competitive image generation under modest hardware constraints~\citep{photoroom2025prx, bhanded2025speedrunning}. In video, efficiency work has more often focused on reducing per-step complexity directly, for example through linear-attention variants as in SANA-Video or aggressive latent compression as in LTX-Video~\citep{chen2025sana, hacohen2024ltx, hacohen2026ltx}. These approaches demonstrate that video efficiency is possible, but they leave open whether image-domain efficiency techniques such as representation alignment and token routing can be composed effectively in text-to-video training.

\section{Model Architecture}
\label{sec:model_architecture}

\begin{figure}[t]
    \centering
    \includegraphics[width=0.8\textwidth]{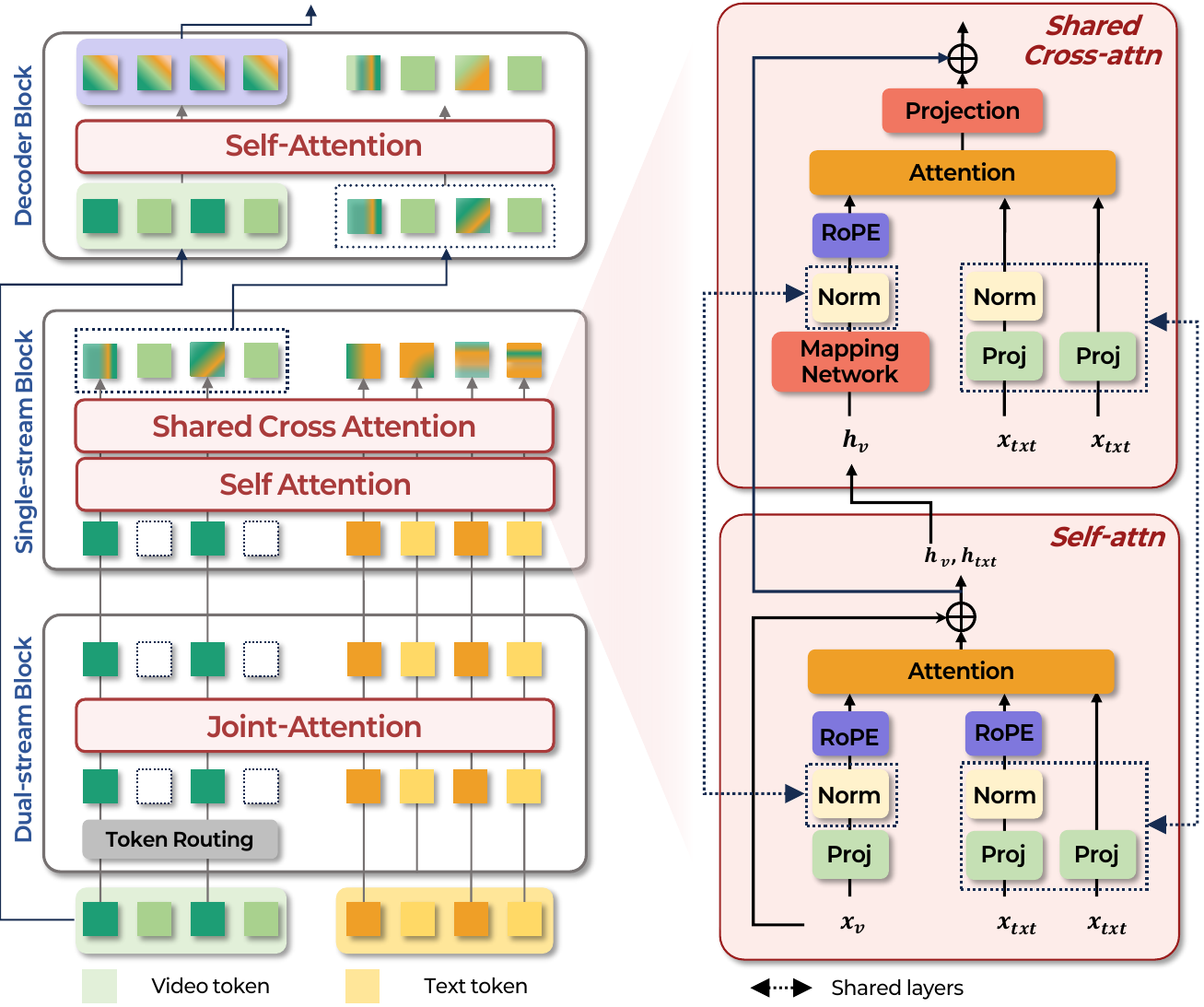}
    \caption{\textbf{Overview of Motif-Video 2B.} Text is encoded by T5Gemma2, while video frames are compressed by the Wan2.1 VAE into spatiotemporal latents and patchified into tokens. The transformer backbone follows a three-stage design that separates early modality fusion, joint text-video representation learning, and final detail reconstruction: 12 dual-stream layers preserve modality-specific processing during early fusion, 16 single-stream layers build a joint text-video representation, and 8 DDT decoder layers serve as a dedicated decoder for high-frequency detail reconstruction. Shared Cross-Attention is attached to the single-stream stage to reinforce text conditioning under long-context token imbalance by using learned query/output projections while reusing the enclosing block's self-attention key and value projections. The denoised latent is then unpatchified and decoded by the VAE to produce the final video.}
    \label{fig:main_fig}
\end{figure}

\subsection{Overview}
\label{sec:arch:overview}
The architecture of Motif-Video 2B is organized around a single principle: each component is assigned a well-defined responsibility, and components with conflicting objectives are not asked to share capacity. Concretely, we separate early modality fusion, joint text-video representation learning, and final detail reconstruction rather than forcing a single block type to optimize all three at once.

Text conditioning is handled by T5Gemma2, a multimodal encoder-decoder language model adapted from Gemma 3 via the UL2 objective~\citep{t5gemma2}. We use an encoder-decoder text encoder deliberately: prior work shows that encoder-decoder architectures retain an advantage in bidirectional contextual representation for visual generation, and that even older T5-family encoders can outperform stronger decoder-only LLMs when used as frozen text encoders~\citep{wang2025comprehensive}. In our setting, T5Gemma2 provides the text representation backbone for all stages of generation.

On the video side, input frames are compressed by the Wan2.1 VAE with 8×8 spatial and 4× temporal compression, then patchified with a 2×2×1 kernel to produce the token sequence entering the transformer. The backbone itself follows a three-stage DDT-style encoder-decoder layout that instantiates the role-separation principle explicitly: 12 dual-stream layers preserve modality-specific processing during early fusion, 16 single-stream layers build a joint text-video representation, and 8 decoder layers separate low-frequency semantic encoding from high-frequency detail reconstruction. Shared Cross-Attention is attached to the single-stream stage to reinforce text conditioning once the token sequence becomes dominated by video patches.

For completeness, the full backbone uses 28 encoder layers and 8 decoder layers with QK-normalization throughout, 12 attention heads of dimension 128, and a hidden dimension of 1536. The denoised latent is then unpatchified and decoded by the VAE to reconstruct the output video. Figure~\ref{fig:main_fig} illustrates the full pipeline.

\subsection{Functional Decomposition of the Backbone: Modality Fusion, Joint Representation, and Decoding}
\label{sec:architecture_stage}
The three-stage layout of Motif-Video 2B reflects a deliberate progression of responsibilities: early layers establish modality-aware representations before fusion, middle layers build joint text-video representations, and final layers decouple semantic structure from detail reconstruction. Each transition is motivated by a distinct objective interference that arises when these responsibilities are conflated.

The first 12 layers operate as dual-stream blocks, processing text and video tokens through separate self-attention pathways before exchanging information via cross-attention. This separation, introduced in FLUX for image generation, prevents premature entanglement between modalities whose statistical properties differ substantially early in the network. We adopt this design unchanged for video, as the same motivation applies: forcing text and video tokens to share attention capacity before either stream has formed coherent representations degrades both. In this stage, the backbone's role is to establish stable modality-specific features before any fully joint representation is formed.

The subsequent 16 layers operate as single-stream blocks, processing the merged joint sequence. At this stage, text and video tokens attend freely to one another, enabling the model to build the shared representations necessary for text-conditioned generation. This stage therefore carries the main burden of cross-modal integration, but, as we discuss in Section~\ref{sec:shared-cross-attention}, it also introduces a text alignment failure mode under long-context generation that requires explicit correction.

\paragraph{Decoupled decoder layers.} The final 8 layers follow the DDT design, functioning as a velocity decoder atop the preceding 28-layer encoder. The DDT encoder-decoder split resolves an optimization conflict inherent to standard diffusion transformers: low-frequency semantic encoding and high-frequency detail decoding impose competing gradient signals when handled by the same modules~\citep{wang2025ddt}. By delegating detail reconstruction to a dedicated decoder, the encoder is free to build semantically coherent representations without being pulled toward high-frequency objectives~\citep{wang2025ddt}.

\begin{figure}[t]
    \centering
    \includegraphics[width=\linewidth]{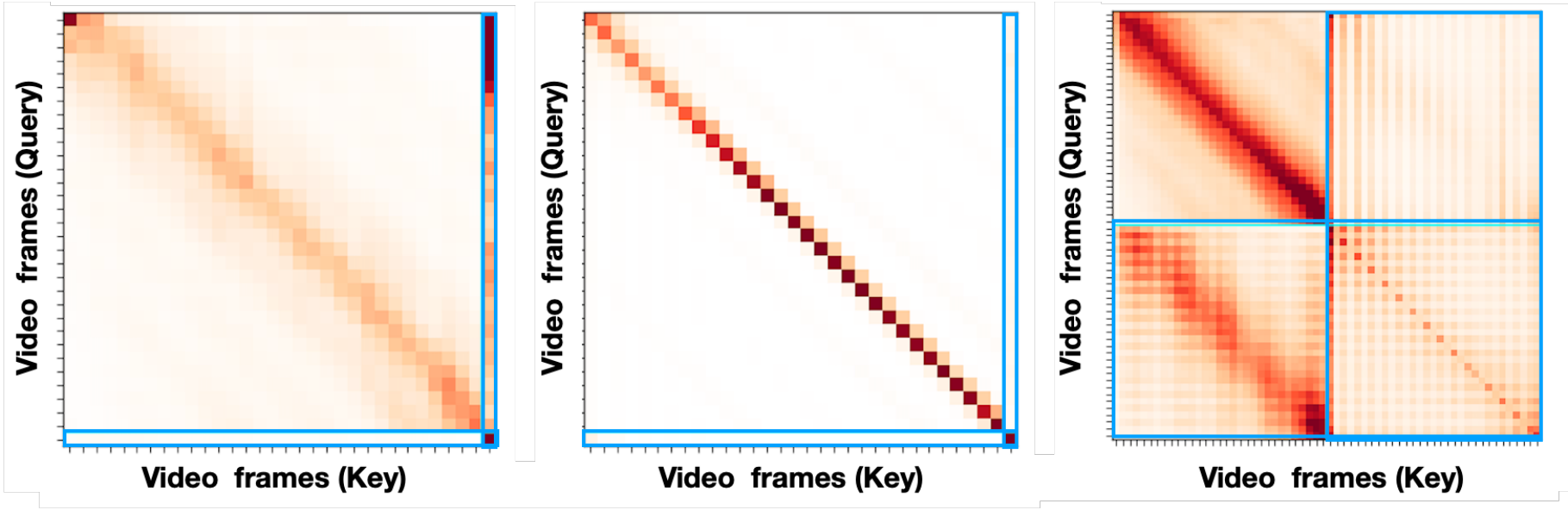}
    \caption{\textbf{Attention structure in dual-stream vs. single-stream vs. DDT decoder layers.} Compared with dual and single-stream layers, DDT decoder layers show stronger inter-frame attention structure, where each frame attends more to temporally adjacent frames. The blue box denotes the encoder hidden state: text tokens in the dual-stream and single-stream cases, and the video output tokens from the encoder layers in the decoder case.}
    \label{fig:ddt-attn}
\end{figure}

In the video setting, this decoupling is associated with an additional effect that we did not anticipate from the image-domain formulation. Attention heatmaps within the DDT decoder blocks reveal a clear inter-frame attention structure, with each frame attending preferentially to temporally adjacent frames rather than distributing attention uniformly across the sequence (Figure~\ref{fig:ddt-attn}). This pattern is present but substantially weaker in the single-stream layers, consistent with the global-attention observations reported by Enhance-A-Video~\citep{luo2025enhance} for other video diffusion transformers, suggesting that the decoupled optimization of the DDT decoder may amplify inter-frame attention as a consequence of its dedicated role. We view this pattern as consistent with an inductive bias toward temporal coherence: once relieved of semantic encoding, the decoder can concentrate more of its attention on resolving fine-grained temporal consistency. Whether this effect is a consequence of the DDT design specifically or of depth alone is a question we leave for future work.

\subsection{Shared Cross-Attention}
\label{sec:shared-cross-attention}

\paragraph{Motivation.}
\textit{How much does a single-stream video transformer actually attend to text?}
The answer, it turns out, depends critically on something as mundane as token count, and the answer is not encouraging.

In single-stream transformer blocks, video and text tokens are concatenated and processed through shared self-attention parameters.
This is elegant: a single pass suffices for cross-modal interaction, and the shared parameterization promotes early alignment.
But elegance conceals a structural problem.
The softmax normalization in attention sums over the \textit{entire} joint sequence.
For a video query token $i$ attending to text token $j$:

\begin{equation}
    \alpha_{ij} = \frac{
        \exp\!\left(q_i^\top k_j / \sqrt{d}\right)
    }{
        \displaystyle
        \sum_{v \in \mathcal{V}} \exp\!\left(q_i^\top k_v / \sqrt{d}\right)
        \;+\;
        \sum_{t \in \mathcal{T}} \exp\!\left(q_i^\top k_t / \sqrt{d}\right)
    }.
    \label{eq:joint_attn}
\end{equation}

Since $|\mathcal{V}| \gg |\mathcal{T}|$, text tokens occupy only a small fraction of the joint sequence, so their aggregate influence on joint-self-attention tends to be relatively diluted as video token count grows. This is a structural consequence of joint-token competition under a shared attention budget, not merely an optimization artifact.

\begin{figure}[t]
    \centering
    \includegraphics[width=\linewidth]{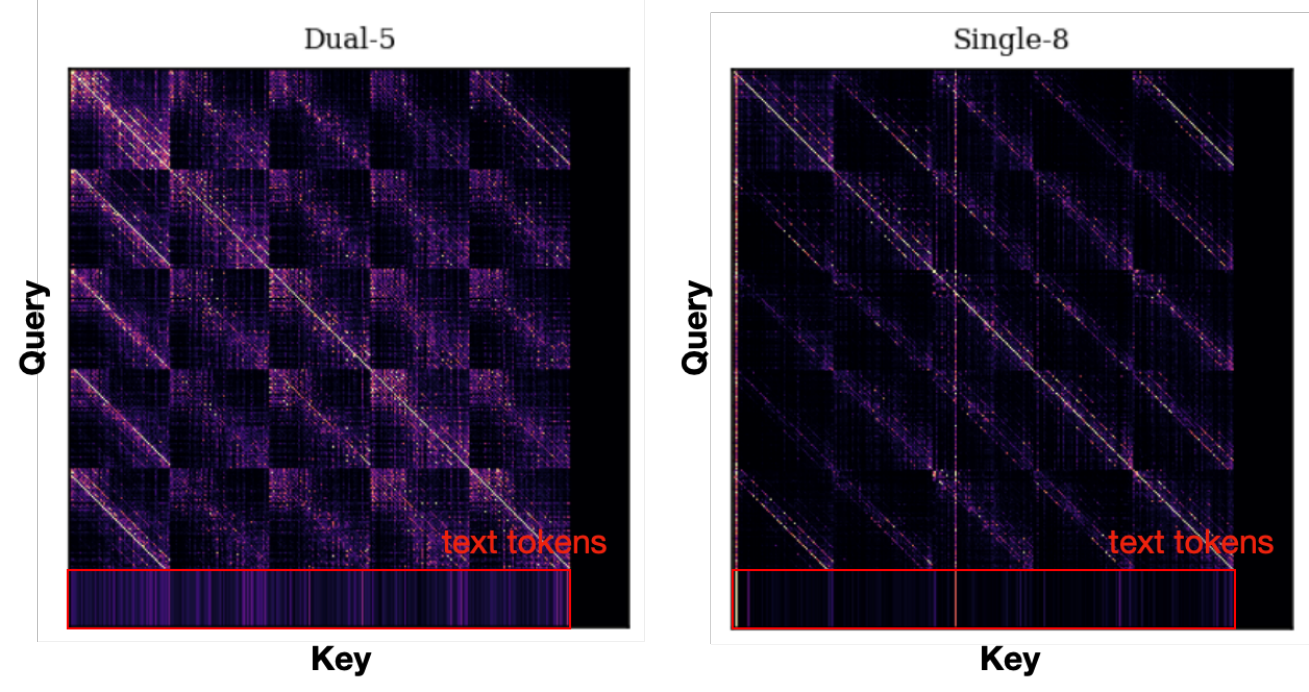}
    \caption{\textbf{Intermediate-layer text-attention drop in single-stream blocks.} We compare attention maps from a representative intermediate layer in dual-stream and single-stream stages. Relative to dual-stream, the single-stream intermediate layer allocates substantially less attention mass to text tokens, indicating weaker text conditioning under joint-token competition.}
    \label{fig:single_layer_attn_drop}
\end{figure}

We confirm this empirically by examining the attention maps of our single-stream transformer blocks (Figure~\ref{fig:single_layer_attn_drop}). In intermediate single-stream layers, the aggregate attention allocated to text tokens is consistently smaller than the attention allocated to video tokens, indicating weaker text influence under joint-token competition.

The structural argument also makes a concrete prediction: as resolution increases, $|\mathcal{V}|$ grows quadratically while $|\mathcal{T}|$ remains fixed, so the dilution should compound with scale.
Indeed, we observe a measurable degradation in prompt following and semantic alignment when scaling training to 720p.
Generated videos exhibit reduced correspondence to fine-grained textual descriptions, a failure mode largely absent at lower resolutions where the $|\mathcal{V}|/|\mathcal{T}|$ imbalance is smaller.
The scaling behavior is precisely what the structural argument predicts.

Together, these two observations, one at the level of attention weights and one at the level of generation quality, point to the same root cause:
joint self-attention, by itself, cannot reliably serve as the sole mechanism for text conditioning in high-resolution single-stream video transformers.
This motivates a dedicated pathway through which text can influence video without competing for a shared attention budget.

\paragraph{Dilution correction is not enough.}
A natural first reaction to the analysis above is to fix the symptom directly: renormalize the attention softmax over text keys alone, removing video tokens from the denominator. This requires no new parameters and is mathematically equivalent to running a second softmax restricted to $\mathcal{T}$. We considered this option and rejected it, because it addresses only the \textit{normalization artifact} and leaves a more fundamental opportunity unused.

The video hidden state $\mathbf{h}_v$ emerging from self-attention is not the same object as the pre-attention input $\mathbf{x}_{\mathrm{pre}}$: it has already aggregated information from neighboring video tokens and formed local spatiotemporal structure that $\mathbf{x}_{\mathrm{pre}}$ did not contain. We would like to ask, conditioned on this newly formed local structure, \textit{which text concepts are now relevant}. A pure renormalization cannot ask this question; it can only re-weight the answers to the question $\mathbf{x}_{\mathrm{pre}}$ already asked. What we want is not a correction to self-attention's output, but a \textit{second, sequential query} into text, posed from the vantage point of what self-attention has just produced.

We refer to this as \textbf{sequential refinement}. It is a strict generalization of dilution correction: any renormalization-only fix is recoverable as a special case in which the refinement query degenerates to the original self-attention query.

\paragraph{Method.}
We append a lightweight cross-attention module to each single-stream transformer block, immediately after self-attention.
Let $\mathbf{h}_v$ denote the self-attention output for video tokens, $\mathbf{x}_{\mathrm{txt}}$ the text hidden states entering the enclosing self-attention layer (i.e., the pre-attention input on the text side),
and $W_K$, $W_V$ the key and value projection weights of that same self-attention layer.
Shared Cross-Attention is defined as:

\begin{align}
    \mathbf{Q} &= W_Q^{\mathrm{cross}}\, \mathbf{h}_v \\
    \mathbf{K} &= W_K\, \mathbf{x}_{\mathrm{txt}}, \qquad
    \mathbf{V}  = W_V\, \mathbf{x}_{\mathrm{txt}} \\
    \mathbf{h}_v &\leftarrow \mathbf{h}_v + W_O^{\mathrm{cross}} \cdot \mathrm{Attn}(\mathbf{Q}, \mathbf{K}, \mathbf{V}),
\end{align}

where $W_Q^{\mathrm{cross}}$ and $W_O^{\mathrm{cross}}$ are the only newly introduced parameters,
$W_K$ and $W_V$ are \textbf{shared} with the enclosing self-attention layer,
and $W_O^{\mathrm{cross}}$ is zero-initialized.
Because $\mathbf{x}_{\mathrm{txt}}$ is precisely the input the enclosing self-attention layer consumes for its own text-side projections, the keys and values produced above are \textit{bitwise identical} to those already computed inside self-attention.
Our implementation reuses the same tensors rather than recomputing them, so the cross-attention adds zero key/value projection FLOPs.


\paragraph{Why $K, V$ are shared but $Q$ is not.}
The design is asymmetric, and the asymmetry is the central point.
$W_K$ and $W_V$ are \textit{content projections}: they map text tokens into a representational subspace that is, by construction, additively compatible with the video residual stream.
Self-attention has spent its training signal arranging exactly this compatibility, since text values already contribute to $\mathbf{h}_v$ as a summand in the joint softmax of Eq.~\eqref{eq:joint_attn}.
Discarding $W_K, W_V$ in favor of freshly initialized cross-attention projections would require relearning this geometric alignment from scratch, and would do so under a much weaker training signal than the joint self-attention provided.
Sharing them is not a parameter-saving trick; it is a commitment to perform the refinement \textit{within the representational manifold the model has already established}.

$W_Q$, in contrast, is a \textit{query projection}: it encodes what the layer is asking about.
Self-attention's $W_Q$ was trained to operate on $\mathbf{x}_{\mathrm{pre}}$, the pre-attention input, and to formulate queries appropriate to that representation.
The refinement target $\mathbf{h}_v$ is a different object: it lives downstream of self-attention and contains aggregated local context that $\mathbf{x}_{\mathrm{pre}}$ did not.
Reusing $W_Q$ on $\mathbf{h}_v$ would amount to asking a question designed for one input distribution from a different one, a quiet but real distribution shift.
More importantly, it would force the refinement query to be the \textit{same} question self-attention already asked, foreclosing the entire purpose of sequential refinement.
We therefore introduce $W_Q^{\mathrm{cross}}$ as a freshly learned projection whose role is to map $\mathbf{h}_v$ into the query space established by the shared $W_K$.

Although $W_Q^{\mathrm{cross}}$ is parametrically free, it is not geometrically free in the way that matters.
What we require is not that the \textit{learned weight} $W_Q^{\mathrm{cross}}$ resemble $W_Q^{\mathrm{SA}}$. Indeed, if it did, the refinement would collapse into the same question self-attention already asked. What we require instead is that the \textit{resulting queries} $\mathbf{q}^{\mathrm{cross}}_i = W_Q^{\mathrm{cross}}\, \mathbf{h}_{v,i}$ form well-conditioned inner products with the shared keys $\mathbf{K} = W_K\, \mathbf{x}_{\mathrm{txt}}$.
Since $\mathbf{K}$ is fixed by sharing, the flow-matching loss can only be reduced by producing queries that yield meaningful attention distributions over this fixed key set; queries that drift off the key manifold yield near-uniform softmax outputs and contribute no useful gradient.
Manifold compatibility is therefore enforced as an \textit{outcome-level} training constraint: the parameters are free, but the only direction in parameter space that reduces loss is the one that keeps queries in conversation with the shared keys.

\paragraph{On the output projection.}
$W_O^{\mathrm{cross}}$ is not shared with the enclosing self-attention's output projection.
By the same logic as above, manifold consistency would in principle argue for sharing it as well.
We prioritize a different consideration: zero-initialization of $W_O^{\mathrm{cross}}$ guarantees that the augmented block is functionally identical to the base block at initialization, so training begins from a well-defined fixed point and the cross-attention contribution grows gradually as the model learns to use it.
A shared, non-zero $W_O$ would forfeit this stability guarantee.
We treat this as a deliberate trade of geometric purity for optimization stability, and leave an experiment isolating the two choices to future work.

\paragraph{Relation to Prior Work.}
SkyReels-V4~\citep{chen2026skyreels} augments single-stream blocks with a cross-attention layer following self-attention and identifies the same dilution problem we describe. Their formulation, $\mathbf{x}^{\prime\prime}_v = \mathbf{x}^{\prime}_v + \mathrm{Attention}(\mathbf{Q}{=}\mathbf{x}^{\prime}_v,\, \mathbf{K}{=}\mathbf{x}_t,\, \mathbf{V}{=}\mathbf{x}_t)$, takes the post-self-attention video state and the raw text input and uses them directly as $Q, K, V$, with no projection on either side. This eliminates the dilution by restricting the softmax to text keys, but it makes no commitment about how the cross-attention should relate to the geometry self-attention has already established.

Shared Cross-Attention takes a different position, and the answer is deliberately asymmetric. On the key/value side, rather than attending against the raw $\mathbf{x}_t$, we attend against $W_K\, \mathbf{x}_{\mathrm{txt}}$ and $W_V\, \mathbf{x}_{\mathrm{txt}}$, the very keys and values self-attention computes for text, reused as identical tensors. The cross-attention therefore operates on top of the text geometry self-attention already uses, rather than on a parallel raw-embedding surface.

On the query side, the substrate is the same as in SkyReels-V4 ($\mathbf{h}_v = \mathbf{x}^{\prime}_v$), but we apply a learnable projection $W_Q^{\mathrm{cross}}$ with QK normalization on top of it, for the sequential refinement reasons argued earlier. The two designs therefore differ on $Q$ and on $K, V$ for distinct reasons: we add learned structure on $Q$ for refinement, and reuse self-attention's existing structure on $K$ and $V$ for stability.

\begin{figure}[t]
    \centering
    \begin{subfigure}[b]{\linewidth}
        \centering
        \includegraphics[width=\linewidth]{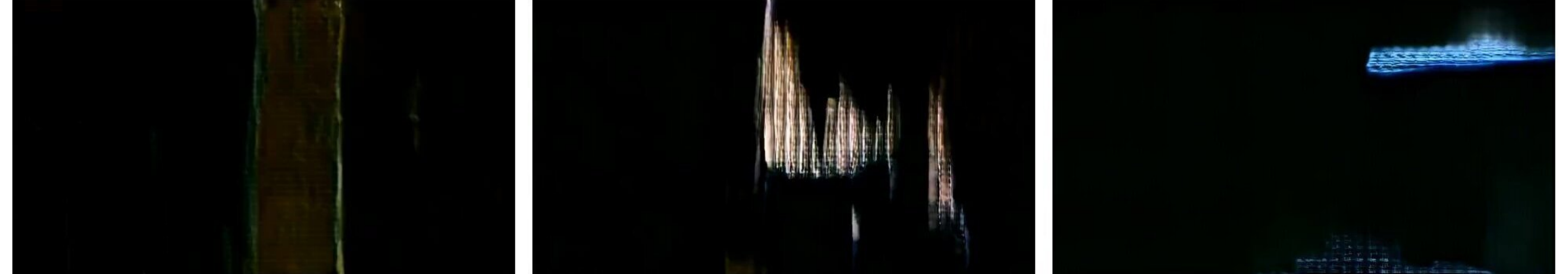}
        \caption{360p continual training w/ SkyReels-V4 cross-attention}
        \label{fig:scattn_vs_skyreels_baseline}
    \end{subfigure}
    
    \vspace{2em}
    
    \begin{subfigure}[b]{\linewidth}
        \centering
        \includegraphics[width=\linewidth]{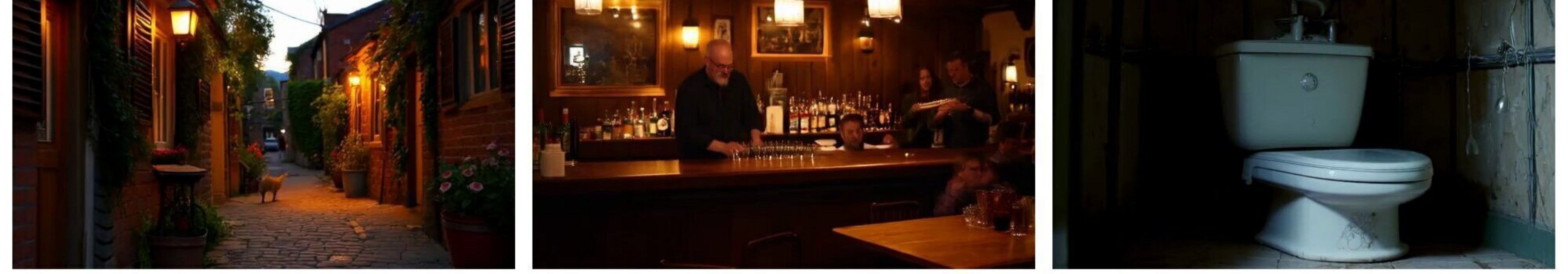}
        \caption{360p continual training w/ shared-cross-attention (ours)}
        \label{fig:scattn_vs_skyreels_ours}
    \end{subfigure}
    \caption{\textbf{Zero-init alone does not save a cross-attention whose $K, V$ geometry is ungrounded.}
    Both variants are inserted into the same pretrained 360p checkpoint with $W_O^{\mathrm{cross}} = 0$, making both forward passes identical to the base model at step 0.
    After 1{,}000 steps of continued training under matched optimizer settings, data, and learning rate, the SkyReels-V4--style cross-attention (\textit{top}) collapses: outputs degenerate to near-black frames with fragmented, incoherent structure, while Shared Cross-Attention (\textit{bottom}) continues training without disruption and produces coherent scenes.
    Each column shows samples from the same prompt under the same seed.
    The contrast directly supports the manifold argument of Section~\ref{sec:shared-cross-attention}: grounding $K, V$ in self-attention's existing projections is what makes a new module stable to insert mid-training.}
    \label{fig:scattn_vs_skyreels_1k}
\end{figure}

\textit{An empirical check.}
The manifold argument makes a falsifiable prediction: if a cross-attention module's $K, V$ have no grounding in self-attention's existing projections, it should fail to integrate stably with an already-trained self-attention pathway, regardless of how carefully it is initialized.
We test this directly.
Starting from the same pretrained checkpoint, we add either the SkyReels-V4--style block or Shared Cross-Attention.
\textit{Both variants zero-initialize their output projection}, so at step 0 each is functionally identical to the base model.
We then continue training for 1{,}000 steps under identical optimizer settings, data, and learning rate.
Figure~\ref{fig:scattn_vs_skyreels_1k} shows the result: the SkyReels-V4--style variant collapses outright, generation degenerates, and the model fails to produce coherent video, while Shared Cross-Attention continues training without disruption.

The mechanism is the one the manifold argument predicts.
Zero-initializing $W_O^{\mathrm{cross}}$ guarantees that the cross-attention contributes nothing at the first forward pass, but it does not freeze the module: gradients still flow through $W_O^{\mathrm{cross}}$, and those gradients depend on $\mathrm{Attn}(Q, K, V)$.
For the SkyReels-V4--style variant, $K$ and $V$ come from raw text embeddings the rest of the network has never been calibrated against, so $\mathrm{Attn}(Q, K, V)$ is essentially noise; the moment $W_O^{\mathrm{cross}}$ becomes nonzero, that noise is injected into the residual stream and propagates through the already-trained self-attention pathway, corrupting downstream representations within a few hundred steps.
For Shared Cross-Attention, $K$ and $V$ are the keys and values self-attention itself uses for text, so the signal $W_O^{\mathrm{cross}}$ learns to inject is small but coherent with the manifold self-attention already operates on.
This is not a claim about the eventual ceiling of either design; a SkyReels-V4--style cross-attention trained from scratch may well learn to recover.
The claim is narrower: when a new module must interface with an already-trained self-attention pathway, grounding its $K, V$ in self-attention's existing projections is what makes that interface stable from the first gradient step.

\section{Training Strategy}
\label{sec:training_strategy}
The architectural choices described in Section 3 define what the model can learn; the training recipe determines whether it actually learns it within a fixed compute budget. For Motif-Video 2B, that budget is tight, roughly an order of magnitude less data and compute than comparably performing open models. Under this constraint, each training iteration must maximize learning efficiency and contribute directly to measurable progress.

Our recipe is built around two ideas. First, we front-load learning by aligning early-stage representations to a frozen visual encoder (REPA with V-JEPA), then remove the alignment objective before it becomes a capacity bottleneck. Second, we treat training as a diagnostic loop rather than a single forward pass through a predefined schedule. When scaling to 720p revealed a regression in semantic alignment, we introduced Shared Cross-Attention mid-training and re-trained at lower resolution before resuming high-resolution adaptation. The remainder of this section describes the full curriculum (Section 4.1), the two efficiency techniques that compose it, representation alignment (Section 4.2) and token routing (Section 4.3), and the iterative refinement process that shaped the final model (Section 4.5).

\subsection{Pre-training and Post-training}

\paragraph{Training objective.}
We train with rectified flow matching~\cite{esser2024scaling, lipman2022flow}. Given a data sample $\mathbf{x}_0$ and noise $\boldsymbol{\epsilon} \sim \mathcal{N}(\mathbf{0}, \mathbf{I})$, the forward interpolation is $\mathbf{x}_t = (1 - t)\mathbf{x}_0 + t\boldsymbol{\epsilon}$ for $t \in [0, 1]$. The model predicts the velocity field $\mathbf{v}_\theta(\mathbf{x}_t, t) \approx \boldsymbol{\epsilon} - \mathbf{x}_0$, and is trained with the standard loss:
\begin{equation}
    \mathcal{L}_{\text{FM}} = \mathbb{E}_{t, \mathbf{x}_0, \boldsymbol{\epsilon}} \left[ \left\| \mathbf{v}_\theta(\mathbf{x}_t, t) - (\boldsymbol{\epsilon} - \mathbf{x}_0) \right\|_2^2 \right].
\end{equation}
We apply classifier-free guidance training with a prompt dropout probability of $p = 0.1$. No modifications are made to the noise schedule or loss weighting; we use the conventional setup throughout.

\paragraph{Image pre-training.}
Training begins with a text-to-image stage at 144p resolution using a sentence-level text embedding model as the conditioning encoder. This stage serves two purposes: it initializes the spatial generation pathway before introducing the complexity of temporal modeling, and it provides a stable starting point for representation alignment with a frozen DINOv2 encoder (Section~\ref{sec:repa}). By decoupling spatial and temporal learning, the model acquires basic compositional and aesthetic capabilities at minimal compute cost before any video data is introduced.

\paragraph{Image--video joint training.}
All subsequent stages train jointly on images and video clips. Image samples stabilize per-frame visual quality and reinforce semantic grounding, while video samples drive temporal modeling. When transitioning from 360p to 480p, we first train on 360p video jointly with 480p images before introducing 480p video. This resolution bridge allows the model to acquire higher-resolution spatial features from images, which are cheaper to process than video, before adapting its temporal pathway to the increased token count. 

\paragraph{Progressive training.}
We increase resolution and frame count in stages, summarized in Table~\ref{tab:training_schedule}. Each transition is made only after the current stage shows diminishing returns on training loss and qualitative evaluation. Inspired by the class-conditioned to text-conditioned curriculum of PixArt-$\alpha$~\cite{chen2023pixart}, we begin training with a sentence-level embedding model and switch to T5Gemma2 at the 360p stage, under the hypothesis that a lower-dimensional conditioning space accelerates early convergence before fine-grained compositional control becomes necessary.

As a rough sanity check on early-stage efficiency, we compared our FID during image pre-training against the compute--performance scaling law of~\cite{liang2024scaling}. At $6.5 \times 10^{20}$~FLOPs, their fitted curve predicts FID~$\approx 30$ for a vanilla DiT, whereas our model reaches FID~$15.5$ under the same budget. The comparison is confounded with concurrent REPA and architectural differences, so we treat it as a consistency check rather than an isolated validation of the early-stage curriculum design.

\begin{table}[h]
\centering
\caption{Simplified training curriculum for Motif-Video 2B. Joint image--video training is used at all video stages. REPA is disabled from Stage~4 onward following evidence that alignment becomes counterproductive after early convergence~\citep{wang2025repa}. Shared Cross-Attention is introduced at Stage~9 to address semantic degradation observed at 720p}
\label{tab:training_schedule}
\small
\begin{tabular}{clccccl}
\toprule
Stage & Task & Resolution & Frames & Text Encoder & REPA & Notes \\
\midrule
1 & T2I & 144p & 1 & Sent.\ Emb. & DINOv2 & Spatial bootstrap \\
2 & T2IV & 144p & 33 & Sent.\ Emb. & V-JEPA & \\
3 & T2IV & 144p & 65 & Sent.\ Emb. & V-JEPA & \\
4 & T2IV & 360p & 65 & T5Gemma2 & -- & REPA disabled \\
5 & T2IV & 480p & 65 & T5Gemma2 & -- & Res.\ bridge first \\
6 & T2IV & 480p & 121 & T5Gemma2 & -- & \\
7 & T2IV & 480p & 121 & T5Gemma2 & -- & SFT \\
8 & T2IV & 720p & 121 & T5Gemma2 & -- & From SFT ckpt \\
9 & T2IV & 360p & 121 & T5Gemma2 & -- & +Motif-cross-attn \\
10 & T2IV & 720p & 121 & T5Gemma2 & -- & Final pretrain + SFT \\
\bottomrule
\end{tabular}
\end{table}

\paragraph{On supervised fine-tuning.}
We perform supervised fine-tuning (SFT) twice during training, at 480p (Stage~7) and 720p (Stage~10), each time on a curated high-quality subset described in Section~\ref{sec:data}. The purpose of SFT is straightforward: it shifts the model's output distribution toward the high-quality tail of the training data, improving aesthetic quality, motion smoothness, and prompt adherence in a way that broad pretraining on loosely filtered data cannot. This follows the now-standard practice in video generation, where HunyuanVideo~1.5~\cite{wu2025hunyuanvideo}, Wan2.1~\cite{wan2025wan}, SANA-Video~\cite{chen2025sana}, Seedance~1.5~\cite{seedance2025seedance}, and SkyReels-V2~\cite{chen2025skyreels} each report a dedicated SFT stage on manually or model-filtered high-quality data after large-scale pretraining.

What is less standard is our choice to initialize the 720p pretraining stage (Stage~8) from the 480p SFT checkpoint rather than from the 480p pretrain checkpoint. The conventional pipeline reserves SFT as a terminal refinement step, applied only after all resolution and frame-count scaling is complete.

We hypothesize that starting from an SFT checkpoint may be preferable when transitioning to a substantially higher resolution: because SFT concentrates the model's learned density on the high-quality manifold, the 720p stage inherits a cleaner starting distribution and can allocate its capacity toward resolution-specific adaptation rather than simultaneously recovering quality lost during broad pretraining. This is analogous to the observation in the LLM post-training literature that each round of alignment produces a better initialization for subsequent training~\cite{grattafiori2024llama}, and to the practice in SkyReels-V2~\cite{chen2025skyreels}, where a 480p SFT checkpoint is used as the starting point for subsequent training stages.

We did not ablate this choice against the alternative of initializing from the pretrain checkpoint. The decision was made early in our training schedule based on the reasoning above, and we observed no instability or regression during the 720p stage. We therefore report it as a pragmatic recipe decision rather than a validated finding, and note it here for reproducibility. The SFT dataset composition and filtering criteria are described in Section~\ref{sec:data}.

\subsection{Representation Alignment (REPA)}
\label{sec:repa}

\paragraph{Background.}
Training diffusion transformers from scratch is expensive partly because the early layers must first discover structured visual representations before the model can make substantial progress on the generation objective. REPA~\cite{yu2024representation} addresses this by adding an auxiliary loss that aligns intermediate DiT hidden states to features from a frozen, pretrained visual encoder. Concretely, let $\mathbf{h}_l$ denote the hidden state at layer $l$ of the DiT, and $\mathbf{z}$ the corresponding feature from the frozen encoder. REPA minimizes:
\begin{equation}
    \mathcal{L}_{\text{REPA}} = - \frac{\mathbf{h}_l \cdot \mathbf{z}}
    {\|\mathbf{h}_l\| \, \|\mathbf{z}\|},
\end{equation}
alongside the primary flow-matching loss $\mathcal{L}_{\text{FM}}$. The alignment target provides a structured learning signal that bypasses the slow self-supervised discovery of spatial structure, accelerating early convergence by over an order of magnitude on ImageNet benchmarks~\cite{yu2024representation}.

\paragraph{Application to video.}
We apply REPA during Stages~1--3 of our training curriculum (Table~\ref{tab:training_schedule}), covering image pre-training and low-resolution video training. We use V-JEPA~\citep{bardes2024revisiting} as the teacher encoder to match the modality. Because V-JEPA learns temporal structure in its latent representations, it is a natural alignment target during the model's initial motion-learning phase.

\paragraph{Phase-constrained alignment.}
We disable REPA from Stage~4 (360p) onward. The rationale follows recent findings on the dynamics of representation alignment during diffusion training: REPA helps in the early phase, when the model's internal representations are still unstructured, but becomes counterproductive once the model's representational capacity exceeds what the frozen teacher can provide~\cite{wang2025repa}. Beyond that point, continued alignment constrains the model to a representational subspace that may not be optimal for the target generation distribution. In our setting, the 360p stage marks the transition from learning global semantics to fine-grained spatial and temporal synthesis, precisely the regime where a frozen teacher is least informative.

\paragraph{On the choice of REPA teacher for video.}
Effective representation alignment depends not only on when to align, but also on what to align to. Recent work by~\cite{singh2025matters} shows that the spatial structure of the teacher's dense features, rather than its global semantic accuracy, is the primary driver of REPA's effectiveness for image generation. That observation matters even more for video, where the teacher must additionally provide temporally coherent spatial structure.

We initially experimented with VideoREPA~\cite{zhang2025videorepa}, which extends REPA to video through Token Relation Distillation (TRD), a second-order objective that aligns pairwise token similarity structure rather than per-token features directly. In our setting, this approach did not yield meaningful improvements on VBench relative to standard per-frame alignment. We suspect two factors contributed. First, TRD transfers relational structure between tokens, but not the dense spatial features themselves, which~\cite{singh2025matters} identifies as the main driver of alignment effectiveness. Second, the underlying teacher, V-JEPA 2.0~\cite{bardes2024revisiting}, provides strong global motion understanding but produces spatially fragmented dense features. That limitation is explicitly identified and addressed by the concurrent V-JEPA 2.1~\cite{mur2026v}, which introduces a dense predictive loss and deep self-supervision to produce spatially structured, temporally consistent representations.

\begin{wrapfigure}[16]{r}{0.40\textwidth}
    \centering
    \includegraphics[width=0.38\textwidth]{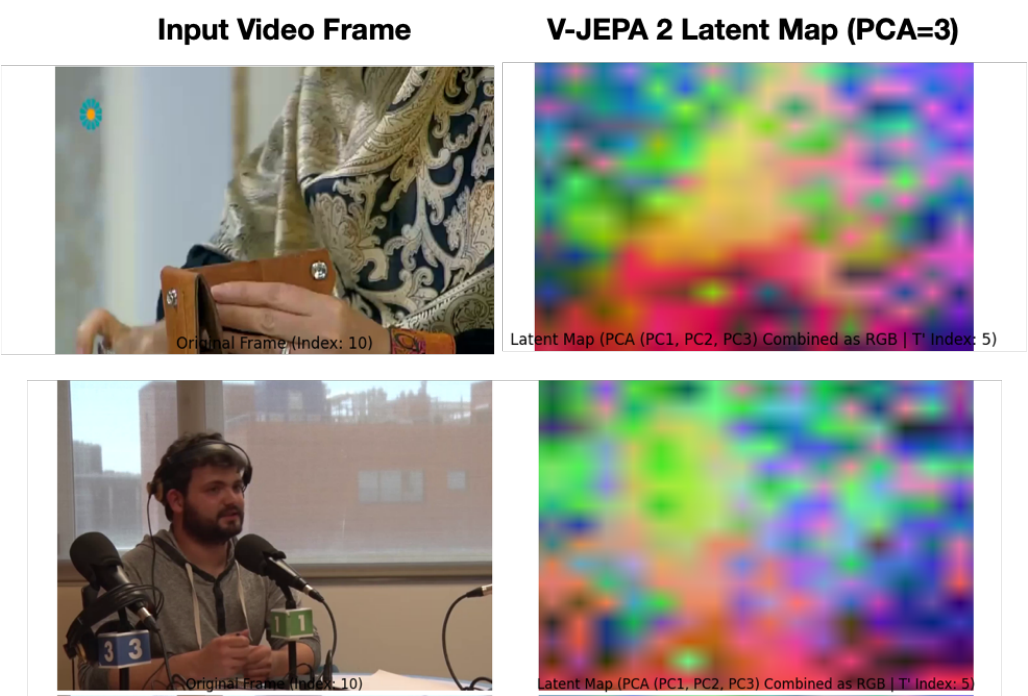}
    \caption{\textbf{Dense features from V-JEPA 2.0.} The visualization highlights that, while V-JEPA 2.0 captures global motion structure well, its dense features are less spatially coherent than would be ideal for dense REPA supervision in video generation.}
    \label{fig:jepa_vis}
\end{wrapfigure}

We include a qualitative comparison of V-JEPA 2.0 dense features in Figure~\ref{fig:jepa_vis}, which illustrates the spatial incoherence that limits its utility as a REPA target.

Taken together, these observations suggest a clear future direction: combine direct dense alignment, as in standard REPA, with a teacher that provides spatially coherent video features, such as V-JEPA 2.1. For the present model, we adopt a pragmatic compromise: we use V-JEPA 2.0 during the early training phases, when global structure dominates, and disable alignment before dense spatial quality becomes the binding constraint.

In practice, we align hidden states from a single intermediate encoder layer (layer~8) to the frozen teacher features. Following iREPA~\cite{singh2025matters}, we use a convolutional projection (a 3$\times$3 Conv2D with spatial normalization) rather than an MLP, because it better preserves spatial structure during projection. The teacher and student feature maps are reshaped into their spatio-temporal layouts and aligned via trilinear interpolation to a common resolution, after which we compute a global cosine similarity loss:
\begin{equation}
    \mathcal{L}_{\text{REPA}} = 1 - \frac{\hat{\mathbf{h}} \cdot 
    \mathbf{z}}{\|\hat{\mathbf{h}}\| \, \|\mathbf{z}\|},
\end{equation}
where $\hat{\mathbf{h}}$ and $\mathbf{z}$ are the flattened spatio-temporal feature volumes from the student projection and frozen teacher, respectively. We set the REPA loss weight $\lambda$ between $0.1$ and $0.5$ across the early training stages based on training-loss dynamics and qualitative evaluation.

\subsection{Token Routing (TREAD)}
\label{sec:tread}

\paragraph{Background.}
In a standard diffusion transformer, every token passes through every layer, so compute cost scales linearly with depth. TREAD~\cite{krause2025tread} starts from the observation that not all tokens require full-depth processing at every training step. During training, a random subset of tokens is routed from an early layer directly to a deeper layer, skipping the intermediate computation. The resulting FLOP reduction is roughly proportional to the fraction of skipped tokens. Crucially, the routed tokens still receive gradient signals from the deep layers they reach, giving the early layers a form of deep supervision that further accelerates convergence. On ImageNet, TREAD achieves up to a 25$\times$ convergence speedup with minimal quality degradation~\cite{krause2025tread}.

\paragraph{Application to video.}
We apply TREAD routing from layer~4 to layer~25 with a token drop ratio of $0.5$, so half of the tokens at each participating layer bypass the intermediate computation. We keep this configuration fixed throughout the T2IV stages once token routing is enabled, rather than tuning it separately for each resolution. We choose a drop ratio of $0.5$ as a conservative operating point: it is large enough to produce meaningful speedups, but not so aggressive that routed tokens dominate the computation or that qualitative regressions become apparent in routine training-time monitoring.

\begin{itemize}
    \item \textbf{Layers 1--3 (dual-stream, excluded):} These layers process text and video tokens in separate streams. Routing across this stage would bypass the modality-specific processing that prevents premature feature entanglement.
    \item \textbf{Layers 4--25 (dual-stream + single-stream, routed):} Once both streams are established, token routing reduces redundant computation while still allowing gradients from the deeper single-stream layers to propagate back through the routed token paths into the earlier stack.
    \item \textbf{Layers 26--36 (DDT decoder, excluded):} The decoder is responsible for high-frequency detail reconstruction. We therefore exclude this stage from routing, since dropping tokens late in the network was empirically more likely to harm fine spatial detail than to produce useful additional savings.
\end{itemize}

At 720p resolution with 121 frames and 512 text tokens, the full transformer requires approximately 4{,}913 TFLOPs per forward pass. With TREAD routing at a 0.5 drop ratio across layers 4--25, this falls to approximately 3{,}563 TFLOPs, a 27.5\% reduction in theoretical FLOPs that corresponds to an estimated $1.38\times$ speedup.

In practice, measured training throughput at 720p increases by $1.31\times$ in videos per second, confirming that most of the theoretical savings translate into wall-clock improvement despite the modest overhead of the routing mechanism. We treat this setting as the main quality--efficiency operating point used in the full recipe; its downstream effect is therefore evaluated through the end-to-end results in Section~\ref{sec:experiments}, rather than through an isolated TREAD-only ablation.

Following the original TREAD formulation~\cite{krause2025tread}, we disable token routing at inference time and use the full model depth for all tokens during generation.

\subsection{Recipe Composition}
\label{sec:recipe_composition}

REPA and TREAD address complementary bottlenecks in compute-constrained training: REPA improves what is learned per iteration by providing a structured alignment target during early training, while TREAD reduces the cost of each iteration by routing redundant tokens past intermediate layers. In our training pipeline, the two techniques therefore operate along different axes of efficiency rather than competing for the same role.

Each component is individually motivated by prior literature, but the point of their combination in our setting is primarily practical: under a fixed compute budget, improving sample efficiency and lowering per-step cost are both necessary to make 2B-scale video training viable.

We do not isolate their individual contributions in this work because our focus is the effectiveness of the full recipe rather than a component-wise ablation study. We therefore evaluate the composition through the end-to-end behavior of the final system in Section~\ref{sec:experiments}, where the relevant question is whether the overall recipe produces a stronger model under the same training budget.

\subsection{Image-to-Video Extension}
\label{sec:i2v}

We train a single model that supports both text-to-video (T2V) and image-to-video (I2V) generation with shared weights. I2V is introduced as an extension of the main T2V training recipe rather than as a separate model family, so the design question is how to use the reference frame strongly enough to preserve subject identity, composition, and appearance without letting it become a shortcut that suppresses motion generation.

Recent I2V systems converge on two complementary observations. First,
first-frame latent conditioning is the most direct way to anchor the generated video
to the input image, because it preserves exact low-level appearance cues.
Second, first-frame latent conditioning alone is often too strong: if the model always sees
a clean first-frame latent, it can learn to preserve the reference appearance by
simply reconstructing or copying from that first frame, instead of learning how
the scene should evolve over time after the first frame.

The first issue is addressed by dual-path conditioning, as in HunyuanVideo 1.5,
which combines latent-level conditioning with image-semantic features~\citep{wu2025hunyuanvideo};
the second is addressed by degrading the conditioning image at high noise levels,
as in Adaptive Low-Pass Guidance, so that motion must be inferred rather than
copied~\citep{choi2025enhancing}. Our implementation follows the same logic,
but adapts it to the Motif-Video backbone and training recipe. The key design choice is to separate exact appearance anchoring, global image semantics, and long-context text alignment into distinct pathways rather than forcing a single conditioning mechanism to solve all three problems.

\subsubsection{Dual Conditioning Pathway}
We condition the model through two complementary pathways: a latent pathway
that anchors exact appearance from the first frame, and a semantic pathway
that supplies a more global image-level summary.

\paragraph{Latent pathway.}
We inject the first frame along a latent pathway for exact appearance
anchoring. Let $\mathbf{I}_1$ denote the first frame of the conditioning video.
Let $E$ denote the Wan2.1 VAE encoder. We first encode this frame into a clean
latent
\begin{equation}
\mathbf{z}_{1} = E(\mathbf{I}_{1})
\in \mathbb{R}^{C \times H \times W},
\end{equation}
with $C=16$ in our setting.

We then construct a conditioning video latent
$\mathbf{z}^{\mathrm{cond}} \in \mathbb{R}^{C \times F \times H \times W}$ by
placing $\mathbf{z}_{1}$ at the first temporal position and zero-filling the
remaining frames:
\begin{equation}
\mathbf{z}^{\mathrm{cond}}(t) =
\begin{cases}
\mathbf{z}_{1}, & t = 1,\\
\mathbf{0}, & t = 2,\dots,F.
\end{cases}
\end{equation}
In parallel, we form a binary mask $\mathbf{m} \in
\mathbb{R}^{1 \times F \times H \times W}$ indicating which temporal positions
are conditioning frames. Let $\mathbf{x}_t \in \mathbb{R}^{C \times F \times H
\times W}$ denote the noisy video latent at diffusion time $t$. The patch
embedding layer receives
\begin{equation}
\mathbf{x}^{\mathrm{in}}_t = \mathrm{Concat}\!\left[\mathbf{x}_t,\ \mathbf{z}^{\mathrm{cond}},\ \mathbf{m}\right],
\end{equation}
which has $16 + 16 + 1 = 33$ input channels. This pathway gives the model
direct access to spatial layout, identity, texture, and color statistics from
the conditioning image.

\paragraph{Semantic pathway.}
On the semantic side, we encode the same first frame into a sequence of image
conditioning tokens. Let $S(\cdot)$ denote the SigLIP vision encoder and let
$P(\cdot)$ denote the lightweight MLP projection. We form
\begin{equation}
\mathbf{s}_{\mathrm{img}} = P\!\left(S(\mathbf{I}_1)\right)
\in \mathbb{R}^{N_{\mathrm{img}} \times D},
\qquad D = 1536.
\end{equation}

These image tokens are then concatenated with the T5Gemma2 text embeddings.
Let $\mathbf{s}_{\mathrm{txt}} \in \mathbb{R}^{N_{\mathrm{txt}} \times D}$ denote the
text-conditioning sequence. The joint conditioning sequence is
\begin{equation}
\mathbf{s}_{\mathrm{joint}} =
\mathrm{Concat}\!\left[\mathbf{s}_{\mathrm{txt}},\ \mathbf{s}_{\mathrm{img}}\right]
\in \mathbb{R}^{(N_{\mathrm{txt}} + N_{\mathrm{img}}) \times D}.
\end{equation}

This mirrors the motivation of recent I2V systems such as HunyuanVideo 1.5~\citep{wu2025hunyuanvideo}: the latent pathway anchors exact appearance, while the image-embedding pathway provides a more global and semantically organized summary that remains useful even when the latent pathway is partially degraded.

The Shared Cross-Attention modules of Section~\ref{sec:shared-cross-attention} operate on the pure T5Gemma2 text embeddings only; SigLIP tokens do not enter the cross-attention context. We keep that separation deliberately: Shared Cross-Attention is introduced to repair text alignment under long video-token sequences, whereas the image embeddings already enter the backbone through the main joint sequence and do not suffer from the same long-context sparsity issue.

\subsubsection{Clean Conditioning Latent with Timestep-Aware Blur}
A second design question is how strongly to expose the first-frame latent during
diffusion training. Injecting the clean conditioning latent unchanged at all
timesteps makes the task too easy in the wrong way: the model can over-rely on
the first frame as a near-copy target. This improves appearance preservation,
but weakens motion synthesis. Adaptive Low-Pass Guidance makes the same
trade-off explicit by degrading the conditioning image more aggressively at high
noise levels and relaxing that degradation as denoising
proceeds~\citep{choi2025enhancing}. We adopt the same core idea, but implement it as
a lightweight timestep-aware blur directly in latent space.

Specifically, we replace the clean first-frame latent $\mathbf{z}_1^{\mathrm{cond}}$ with
\begin{equation}
\tilde{\mathbf{z}}_1^{\mathrm{cond}} \;=\; \mathrm{GaussianBlur2D}\!\left(\mathbf{z}_1^{\mathrm{cond}};\, \sigma(t)\right),
\qquad \sigma(t) = r_{\max} \cdot t,
\end{equation}
where $t \in [0,1]$ is the diffusion timestep and $r_{\max}$ is a fixed maximum blur radius. This linear schedule is a pragmatic choice rather than an ablated optimum: at high noise levels ($t \approx 1$), the conditioning signal is maximally blurred, forcing the model to rely more on text, image semantics, and learned motion priors than on sharp spatial copying. At low noise levels ($t \approx 0$), the blur vanishes and the first-frame appearance is recovered, restoring precise identity and texture control near the end of denoising.

The goal is therefore not to weaken conditioning overall, but to change its
role over the course of denoising. Early in denoising, conditioning should act
as a coarse appearance anchor rather than an exact reconstruction target. Late
in denoising, it should again provide fine-grained appearance fidelity.

\subsubsection{Joint T2V/I2V Training}
A single set of weights handles both T2V and I2V. Once I2V training is enabled, we mix it into the later T2IV stages rather than running a separate I2V-only phase. At each training step, we sample a Bernoulli variable with $p_{\mathrm{i2v}}{=}0.3$ at the batch level, synchronized across FSDP ranks, to decide whether the batch is T2V or I2V. We choose $p_{\mathrm{i2v}}{=}0.3$ as a pragmatic balance: it is large enough for the model to learn stable first-frame conditioning behavior, but small enough to preserve the broader motion prior learned from the dominant T2V batches.

When the batch is I2V, the conditioning pathway described above is activated and a motion-focused caption variant (\texttt{caption\_i2v}) is sampled. T2V batches instead sample among the three caption variants described in Section~\ref{sec:data:captioning}. For classifier-free guidance training, we apply independent dropout with $p{=}0.1$ to both text prompts and SigLIP image embeddings.

We do not introduce a learnable task-type embedding to distinguish T2V from I2V. In practice, the task identity is already explicit in the input: I2V batches contain a non-zero conditioning latent and mask, whereas T2V batches do not. That signal is sufficient for the patch embedding layer, and the caption distribution switch provides an additional cue at the conditioning level. We therefore treat the absence of a task embedding as a deliberate simplification of the recipe, rather than as a separately validated claim. This joint training strategy preserves the broader motion prior learned from pure T2V data, while I2V batches teach the model to anchor that prior to a specific input frame without collapsing into static reconstruction. We evaluate I2V behavior through the end-to-end results rather than through a dedicated ablation of these conditioning choices.

\subsection{Distributed Training}
\label{sec:train:distributed}
We train on 8 Azure nodes, each with 8 H200 GPUs, for a total of 64 GPUs. Jobs are orchestrated with Kubernetes and launched through SkyPilot~\citep{yang2023skypilot}, which handles scheduling, fault recovery, and cloud resource provisioning. This setup lets us treat the Azure cluster as a single training pool rather than managing nodes individually. We use FSDP2 through Accelerate~\citep{gugger2022accelerate}. At the 2B parameter scale of Motif-Video, a single intra-node shard group is sufficient to fit the full model state for our 720p, 121-frame configuration, so we do not require tensor or sequence parallelism. Avoiding those additional parallelism modes simplifies the communication pattern and reduces synchronization overhead.

\paragraph{Sharding strategy.}
We adopt Hybrid Sharded Data Parallelism (HSDP): parameters are sharded across the 8 GPUs within each node and replicated across the 8 nodes ($\text{DP-replica} = 8$). The forward all-gather that materializes full parameters remains within a node over NVLink, keeping the latency-sensitive path off the inter-node network. Across nodes, only the post-reduce-scatter gradient shard, rather than the full parameter tensor, is communicated. In practice, this design provides enough memory headroom for the full 720p configuration without requiring a more complex parallel decomposition.

\paragraph{Activation checkpointing, compilation, and FSDP wrapping order.}
We apply activation checkpointing, \texttt{torch.compile}, and FSDP2 \texttt{fully\_shard} in that order. In our implementation, this requires a small patch to Accelerate's default FSDP2 path. Without that change, checkpointed transformer blocks are not compiled and sharded at the intended granularity. That in turn breaks the block-level wrapping scheme used by our model. Full implementation details are provided in Appendix~\ref{app:fsdp2_impl}.

Training uses \texttt{bfloat16} mixed precision for model computation and activations, while reduction-sensitive communication and optimizer states remain in \texttt{float32}. This configuration preserves the throughput advantage of \texttt{bfloat16} while keeping numerically sensitive reductions and optimizer updates in higher precision.


\section{Data}
\label{sec:data}

\subsection{Data processing pipeline}
\label{sec:data:preprocessing}
Our training corpus combines two sources: an internal web-scale video collection and a set of publicly available video datasets. Rather than maximizing raw scale, we prioritize curation quality to support resource-efficient training, organizing the raw pool into real and synthetic branches for both images and videos and routing each surviving clip through a progressive multi-resolution training schedule. We made extensive use of NeMo Curator~\citep{jennings2024nemo}, whose scalable data-curation toolkit and support for large-scale video-processing pipelines substantially streamlined our preprocessing workflow.

\begin{figure}[t]
    \centering
    \includegraphics[width=\linewidth]{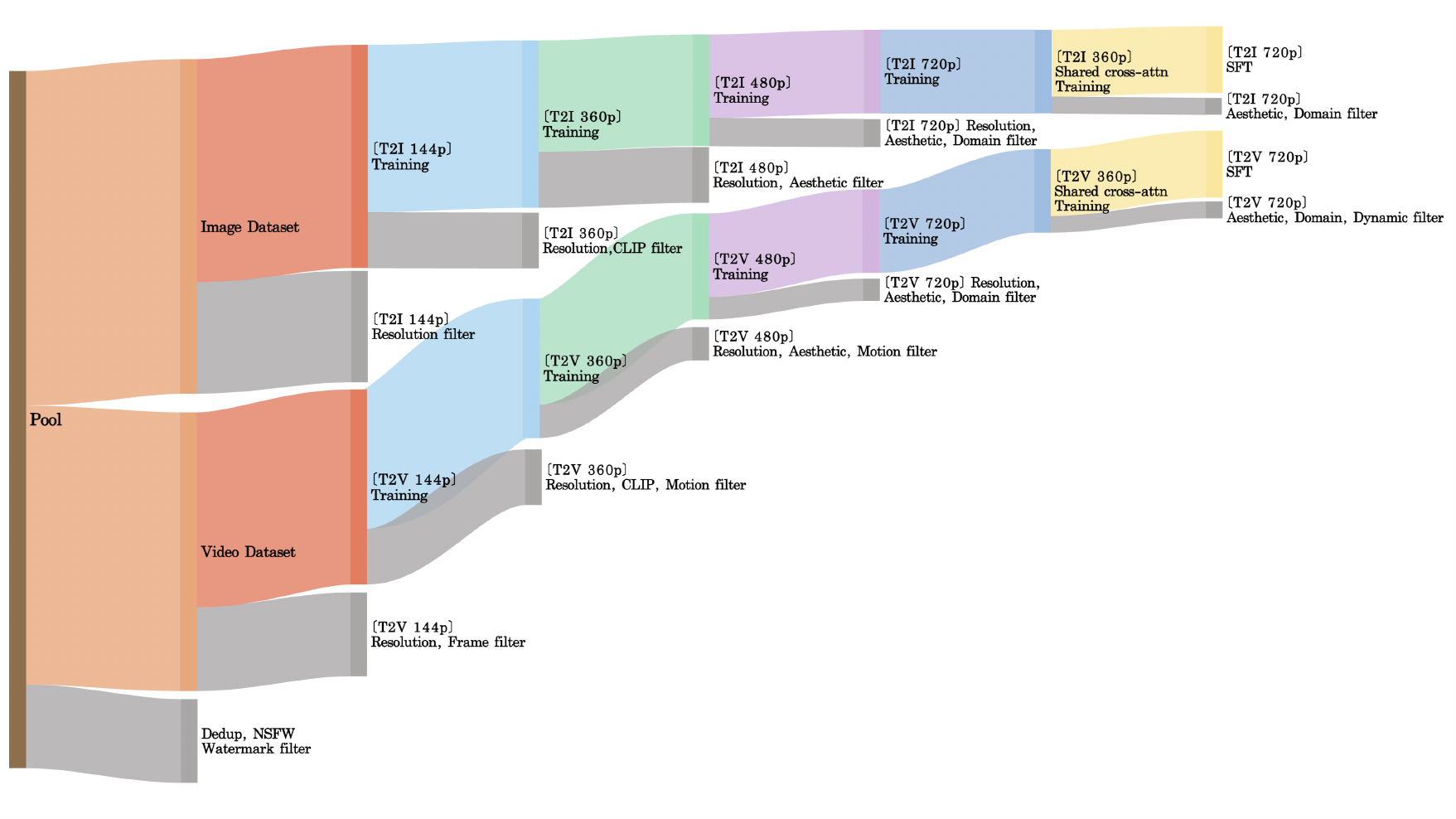}
    \caption{\textbf{Overview of the training-data construction pipeline.} The raw pool is split into Image Real, Image Synthetic, Video Real, and Video Synthetic branches. An initial sanitation stage removes broken files, abnormally small files, near-duplicates (SSCD-based), NSFW content, and watermarked content. Surviving clips are progressively filtered by resolution, clip length, motion, and aesthetic signals as they advance through the 144p, 360p, 480p, and 720p training stages, and by stricter aesthetic, domain, and dynamic-motion criteria before the cross-attention refinement and final 720p SFT stage. The Sankey diagram visualizes how flows contract from the raw pool toward the curated training and SFT corpora.}
    \label{fig:data-pipeline-sankey}
\end{figure}

\subsubsection{Data collecting and preprocessing}

Our training corpus combines an internal web-scale crawl with publicly available image and video datasets. We process both sources through the same downstream pipeline so that the final corpus is governed by a single set of sanitation, filtering, deduplication, and stage-wise quality controls.

\paragraph{Sanitation.}
Before any stage-specific filtering, every raw clip passes through a sanitation block that removes broken or non-decodable files, abnormally small files that typically correspond to thumbnails or corrupted downloads, near-duplicates identified by our SSCD-based deduplication pipeline (described below), NSFW content, and watermarked content.

The NSFW and watermark filters combine two signals. An initial OCR-based screen, inherited from the legacy internal crawling pipeline, flags overlaid channel logos, burned-in subtitles, and other high-confidence watermarks using on-frame text detection. Clips that survive this screen are then re-examined by a vision-language model (see Section~\ref{sec:data:captioning}), which produces structured per-clip tags including \texttt{watermark}, \texttt{nsfw}, \texttt{padded}, \texttt{multi\_scene}, \texttt{timelapse}, and overall \texttt{quality}. Clips whose VLM tags flag any of these attributes are dropped. This second pass acts as a semantically aware safety net on top of OCR. Because this VLM pass is shared with caption generation (Section~\ref{sec:data:captioning}), the filter tags and training captions come from the same forward pass.

\paragraph{Black-bar detection.}
Web-crawled video frequently contains letterbox or pillarbox padding
from mismatched aspect ratios.  We detect these regions using
\texttt{ffmpeg}'s \texttt{cropdetect} filter, which estimates the
maximal content rectangle via luminance statistics, and pass the
resulting crop prior to the downstream encode step.
 
\paragraph{OCR detection.}
Burned-in text, including channel logos, persistent subtitles, and promotional
overlays, cannot be caught by \texttt{cropdetect} alone.  We run
PaddleOCR-VL~\cite{cui2025paddleocr} (served via vLLM) on $N$ uniformly sampled frames per
clip, then cluster detections across frames by spatial IoU and retain
only clusters present in $\geq$50\% of frames.  This persistent-region
filter distinguishes fixed overlays from transient in-scene text.
The surviving OCR regions are composed with the black-bar crop into a
single final rectangle by excluding detections in the top 20\% (logos)
or bottom 20\% (subtitles) of the content area, and the result is
applied in one \texttt{ffmpeg} re-encode pass alongside resolution
scaling and frame-rate limiting.

\paragraph{Scene segmentation and length control.}
For video branches we first detect scene boundaries using a conservative threshold that prefers over-segmentation (false positives) over missed transitions (false negatives), and then merge adjacent segments using stitch detection based on SigLIP embedding similarity, which recovers contiguous shots that were split by momentary motion or exposure changes. Clips shorter than two seconds after merging are discarded to guarantee that every training clip covers a meaningful temporal extent.

\subsubsection{Vision quality filtering and deduplication}

We apply a multi-stage video quality filtering pipeline that scores each sample from complementary perspectives: aesthetic quality, luminance, model-based training suitability, technical quality, and motion quality. These signals are not used as a single learned ranking. Instead, each filter removes a specific failure mode, such as poor exposure, severe compression artifacts, static clips, or temporally unstable motion, before the surviving clips are routed to later training stages.

\paragraph{Aesthetic Quality.}
We assess aesthetic quality using Aesthetic Predictor V2.5~\cite{aes2_5}, a SigLIP-based predictor~\cite{zhai2023sigmoid} that estimates image-level aesthetic scores. For each video, we uniformly sample frames over time, compute frame-wise aesthetic scores, and aggregate them into a single video-level score by averaging across the sampled frames. This score is used as a stage-wise filter: clips in the low-aesthetic tail are removed, and the cutoff becomes stricter at higher-resolution stages.

\paragraph{Luminance.}
Following the formulation adopted in OpenHumanVid, luminance is computed as
\begin{equation}
L = 0.2126\,R + 0.7152\,G + 0.0722\,B,
\end{equation}
where $R$, $G$, and $B$ denote the pixel intensities of the red, green, and blue channels, respectively. We compute luminance statistics over sampled frames and remove videos that fall into the extreme low- or high-luminance tails for the target stage. This procedure filters out severely underexposed or overexposed videos and improves the visibility of subjects and scene content in the retained dataset.

\paragraph{Model-based Suitability Score.}
In addition to low-level visual cues, we incorporate a model-based suitability signal inspired by Koala-36M~\cite{wang2025koala}. This score summarizes multiple quality-related factors into a single estimate of whether a video is suitable for training a video generation model. In practice, we use it conservatively as a rejection filter: clips in the lowest-suitability tail are removed, while the rest remain subject to the other specialized filters below.

\paragraph{Technical Quality.}
We further evaluate the overall technical quality of each video using DOVER~\cite{wu2023exploring}, a video quality assessment model designed to disentangle technical and aesthetic aspects of video quality. In our pipeline, we use the technical-quality-related output to filter out videos affected by compression artifacts, noise, distortion, low sharpness, or other degradations that may negatively affect model training. This step improves the low-level fidelity of the retained videos and reduces noise in the training distribution.

\paragraph{Motion Quality.}
We assess motion quality using optical flow statistics. Specifically, UniMatch~\cite{yang2023revisiting} is employed to estimate optical flow between sampled frame pairs and compute a motion score for each video. We remove both tails of this distribution: extremely low-motion clips are typically static or nearly static, while extremely high-motion clips often contain cuts, jitter, or unstable camera motion. The retained middle band better matches the smooth temporal dynamics targeted by the main training stages.

\paragraph{Progressive stage-wise filtering.}
Surviving clips are routed through a progressive multi-resolution training schedule that alternates image (T2I) and video (T2V) stages at 144p, 360p, 480p, and 720p, each with tighter admission criteria (Figure~\ref{fig:data-pipeline-sankey}). At every transition we re-apply resolution, clip-length, motion, and aesthetic filters, with stricter cutoffs at higher resolutions, so that later stages are trained only on clips that satisfy stronger visual and temporal quality requirements. The final 720p SFT stage adds domain-balancing and, for video, dynamic-motion criteria. Before that final stage, we also run a 360p \emph{Shared Cross-Attention} refinement stage on an already-filtered subset. Synthetic video is injected only at 720p, where its controlled quality is most compatible with the admission criteria.

\paragraph{SSCD-based deduplication.}
We deduplicate the corpus with a three-stage SSCD pipeline~\citep{pizzi2022self}.

\emph{Embedding.} We encode each image or video with the publicly released \texttt{sscd\_disc\_mixup} TorchScript model, producing a 512-dimensional descriptor per frame after resizing to $320{\times}320$ and applying ImageNet normalization. For videos, we use the tenth frame as a representative frame. This choice avoids intro and logo bias from the earliest frames and keeps matching tractable by avoiding all-pairs frame comparison. We use SSCD because it is designed for copy detection and is robust to re-encoding, cropping, and light editing, which are common duplication modes in web-crawled video.

\emph{Grouping.} We search the descriptor set with NVIDIA cuVS's multi-GPU IVF-PQ index under cosine distance~\citep{cuvs}. We retrieve $k{=}64$ neighbors per query with \texttt{nprobe}${=}16$ and keep only pairs whose cosine similarity exceeds $0.9$. We then merge the retained pairs with Union-Find to form duplicate groups.

\emph{Representative selection.} Within each duplicate group we keep a single sample using the weighted score
\[
  s = 0.5\cdot\widehat{\text{res}} + 0.3\cdot\widehat{\text{fps}} + 0.2\cdot\widehat{\text{filesize}},
\]
where each term is min-max normalized inside the group. The remaining members of the group are dropped. This rule favors higher-resolution, higher-frame-rate, and less re-compressed copies.

\subsection{Video captioning}
\label{sec:data:captioning}

\paragraph{Caption-as-metadata.}
Rather than treating captioning as a standalone text-generation step, we use a single vision-language forward pass that returns both natural-language captions and a structured set of downstream-usable tags. All captions and tags in Section~\ref{sec:data:preprocessing} come from Qwen3-VL-30B-A3B~\citep{bai2025qwen3}. For videos, we feed the model $N$ uniformly sampled frames from the clip; for images, we feed the image directly.

We require every response to follow a fixed JSON schema with both free-text and structured fields. In practice, each response contains caption fields together with tags such as \texttt{subject}, \texttt{style}, \texttt{action}, \texttt{camera\_move}, \texttt{quality}, \texttt{watermark}, and \texttt{nsfw}. This \emph{caption-as-metadata} design lets us reuse the same forward pass for (i) text-conditioning during training, (ii) sanitation (\texttt{nsfw}, \texttt{watermark}, \texttt{padded}, \texttt{multi\_scene}, \texttt{timelapse}, \texttt{quality}), (iii) domain- and subject-balanced sampling, and (iv) dynamic-motion filtering at 720p SFT.

\paragraph{Prompt design.}
We use two prompts that share a common JSON schema but differ in their temporal fields. The video prompt treats the sampled frames as a single description target and asks for, in order, camera attributes (shot type, angle, motion), subjects, actions, environment, lighting and color, and any on-screen text. The image prompt removes the temporal fields and instead asks for composition, framing, and verbatim text transcription. In both cases, the schema includes free-text caption fields together with structured fields such as \texttt{style}, \texttt{subject}, \texttt{action}, \texttt{camera\_move}, and \texttt{quality}.

Both prompts forbid claims that are not grounded in the visible frames, frame-by-frame narration, and subjective comments on quality, smoothness, or atmosphere. These constraints are intended to reduce hallucinated tags or descriptive drift. We require each response to be a single valid JSON object; malformed responses are re-sampled.

\paragraph{Caption variants for text-robust training.}
For each clip we retain three caption variants derived from the same VLM response: \texttt{caption\_long} (a detailed 150 to 250 word description), \texttt{caption\_short} (a single 15 to 25 word sentence), and \texttt{caption\_truncated}, obtained by keeping only the leading sentence of \texttt{caption\_long}. During training we sample among the three with fixed probabilities $(p_{\text{long}}, p_{\text{short}}, p_{\text{truncated}}) = (0.5, 0.3, 0.2)$. The intent is to reduce the train--test mismatch between long synthetic captions and the shorter prompts users typically provide at inference time. This is a pragmatic recipe choice rather than an isolated claim of optimality. Short and truncated variants also act as a mild form of caption dropout that reduces overfitting to VLM-specific phrasing.

\paragraph{Filter integration.}
The structured fields produced by the VLM are consumed directly by the sanitation and stage-wise filtering steps of Section~\ref{sec:data:preprocessing}; we do not apply a separate post-processing model to reinterpret them. Specifically, \texttt{watermark}, \texttt{nsfw}, and \texttt{padded} flags trigger hard removal; \texttt{multi\_scene} clips are dropped as a secondary check on scene segmentation; \texttt{quality}${=}$\texttt{low} is excluded from 480p and above; \texttt{style} and \texttt{subject} drive domain balancing for the 720p stage and SFT; and \texttt{action}${=}$\texttt{Dynamic} is used as the dynamic-motion criterion for 720p SFT admission. Because these tags are produced in the same forward pass as the training captions, filtering and conditioning remain synchronized by construction throughout the data pipeline.

\begin{figure}[t]
    \centering
    \includegraphics[width=\linewidth]{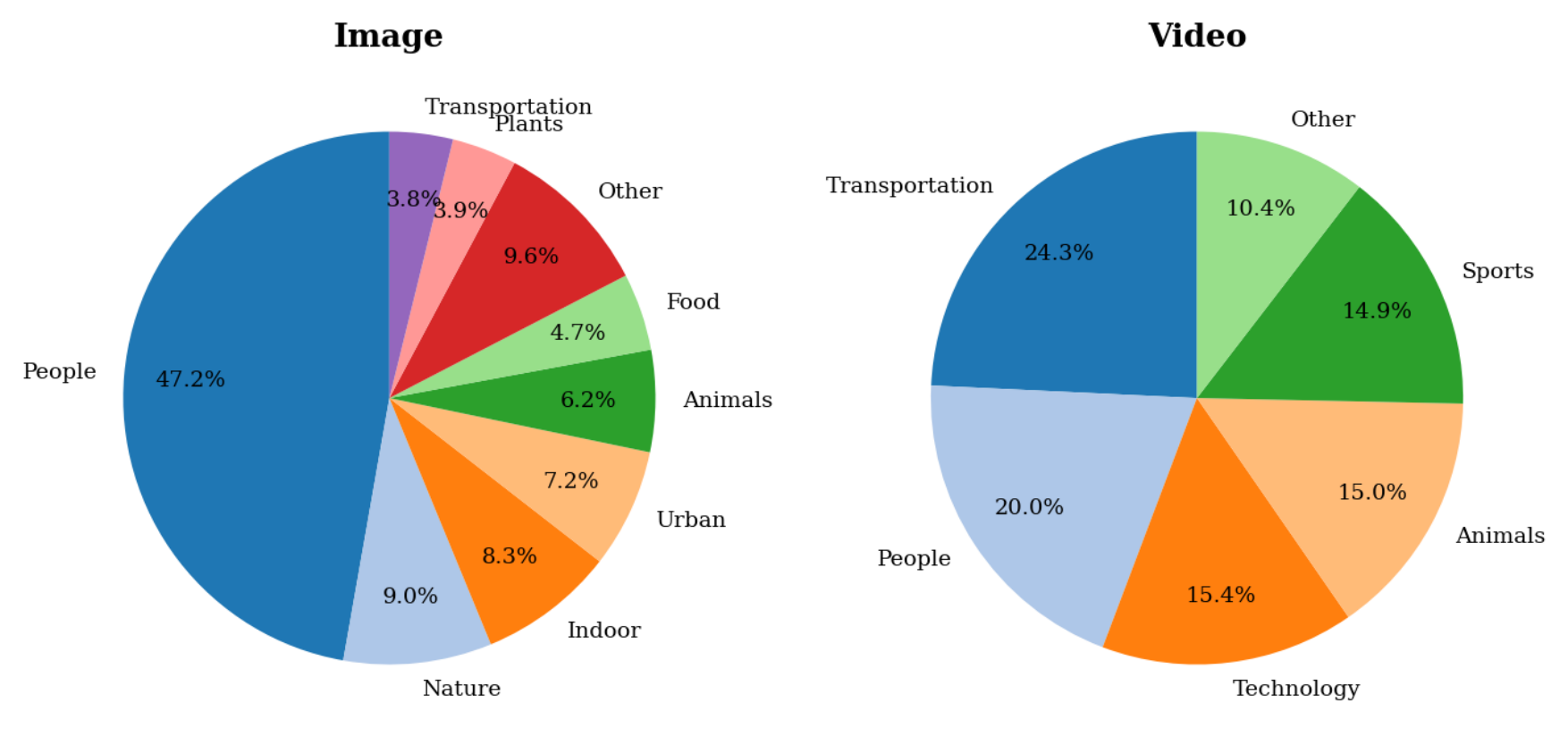}
    \caption{\textbf{Subject composition of the cross-attention fine-tuning corpus.} The corpus was assembled iteratively by curating additional clips from underperforming categories. Left: image distribution. Right: video distribution.}
    \label{fig:sft-corpus-composition}
\end{figure}

\paragraph{Fine-tuning corpus composition.}
As a downstream use of caption metadata, we assembled the fine-tuning corpus (Table~\ref{tab:training_schedule}, Stages~9--10) iteratively. We ran intermediate evaluations on the latest checkpoint, identified subject categories where generation quality was weakest, and then curated additional clips from those categories. Figure~\ref{fig:sft-corpus-composition} shows the resulting subject distribution. For images, \texttt{People} dominates, reflecting character-centric use cases. For videos, the distribution shifts toward \texttt{Transportation}, \texttt{Sports}, and \texttt{Animals}, categories involving dynamic motion that were identified as weak points in intermediate evaluations.

\subsection{Offline Bucket-Balanced Sampler}
\label{sec:data:dbps}
\paragraph{Problem.}
A common storage format for large-scale training is WebDataset, which packs samples into tar shards and supports efficient sequential streaming~\citep{webdataset}.
In our setting, however, training Motif-Video 2B on $W$ GPUs is bottlenecked by data heterogeneity.
Samples vary along three axes, frame count, height, and width, and we must preserve sample-level filtering and bucketed batching without giving up the benefits of shard-based storage.
In practice, we group samples jointly by frame bucket and resolution bucket.
The frame buckets comprise single-frame images, videos with 33, 65 and 121 frames, each of which is further split across multiple spatial resolutions.
Under our FSDP2/HSDP training setup (Section~\ref{sec:train:distributed}), progress on bucket $b$ can proceed only when \emph{all} participating ranks have accumulated a full batch for that same frame-and-resolution bucket:
\begin{equation}
  \text{global steps}^{(b)} = \min_{r}\bigl(\text{steps}_{r}^{(b)}\bigr)
  \quad \forall\, b \in \text{active buckets.}
  \label{eq:dist-min}
\end{equation}
As a result, progress on each active bucket could be limited by the slowest participating rank.
The baseline sampler materializes a global clip index over all shards, applies random shuffling (\texttt{shuffle\_block\_size=1}) that destroys archive locality, and distributes indices by round-robin assignment ($\text{index}\;k \to \text{rank}\;k \bmod W$).
This preserves stochasticity, but it sacrifices WebDataset's main advantage, fast sequential shard reads, and creates substantial cross-rank imbalance in bucket composition.
If a single rank receives too few samples for one frame-and-resolution bucket, updates for that bucket are delayed for the synchronized FSDP2 job, reducing effective utilization across all $W$ GPUs.
Empirically, the baseline yields $N$ steps per epoch at roughly 20\% utilization, with the remaining budget lost to synchronization overhead.
Randomized I/O increases dataloader latency to 0.05\,s/step.

\paragraph{Method.}
Our offline bucket-balanced sampler leaves the underlying WebDataset shard layout unchanged and moves filtering, bucketing, and rank assignment into an \emph{offline planning phase}.
The key idea is to make all expensive selection decisions from metadata offline and then execute the resulting plan with sequential shard reads during training.

\textit{(i) Metadata-driven shard planning.}
Given clip-level Parquet metadata, we first apply filtering rules and assign each surviving clip to a joint frame-and-resolution bucket.
We then build an initial greedy shard assignment by iteratively placing each tar shard on the rank that most reduces the current cross-rank bucket imbalance.
Starting from this greedy initialization, we run a simulated annealing (SA) optimizer~\citep{kirkpatrick1983optimization} for 30{,}000 iterations to refine a shard-to-rank assignment map $\sigma$ that minimizes the coefficient of variation (CV) of 1f, 33f, 65f and 121f clip counts across ranks:
\begin{equation}
  \min_{\sigma}\;\mathrm{CV}\!\left(\bigl\{n_{r,b}\bigr\}_{r=0}^{W-1}\right)
  \quad \forall\, b,
  \label{eq:sa-obj}
\end{equation}
where $n_{r,b}$ denotes the number of bucket-$b$ clips assigned to rank $r$.
Each SA iteration proposes a swap of two tar shards between ranks.
The final assignment is serialized into per-rank shard files (\texttt{rank}\{r\}\texttt{.npz}), each containing an ordered plan over shards and samples.

\textit{(ii) Sequential WebDataset reads.}
At runtime, each rank reads only its assigned tar shards in shard order, so filtering and bucketing no longer require global online reshuffling.
This preserves sequential WebDataset I/O and reduces dataloader latency to below 0.001\,s/step.
A locality-preserving rolling shuffle with a 4\,096-sample window preserves within-bucket randomness without breaking read locality.

\textit{(iii) Image/video interleaving.}
We use a fixed image--video interleaving schedule derived from the planned per-bucket step counts. An example pattern is \texttt{I-V-V}, although the full schedule is determined by the planned bucket counts. This keeps the image/video mixture stable throughout training.

\begin{figure*}[t]
  \centering
  \includegraphics[width=\textwidth]{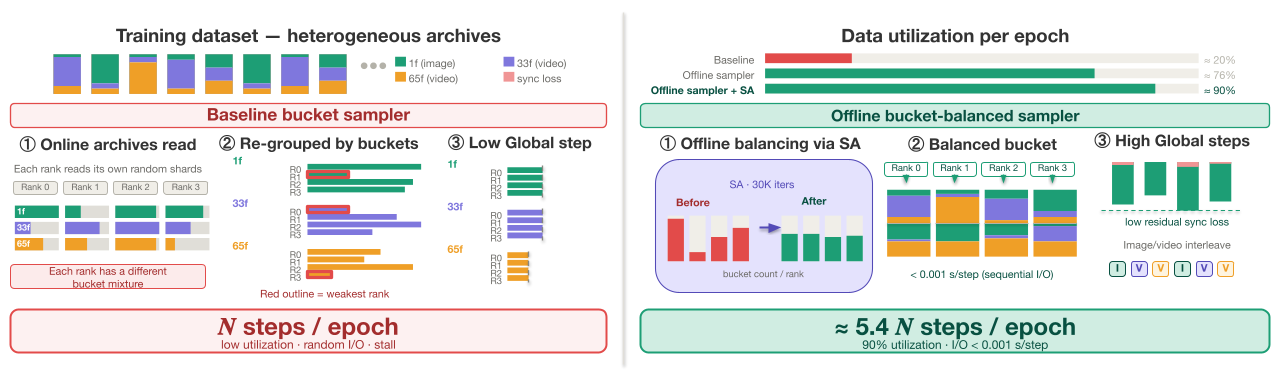}
  \caption{Overview of our offline bucket-balanced sampler for WebDataset-formatted video corpora on $W$ GPUs. An offline planner consumes clip metadata to apply filtering and frame-resolution bucketing, assigns tar shards to ranks, and emits per-rank schedules that preserve sequential archive reads during training.}
  \label{fig:dbps-overview}
\end{figure*}

\paragraph{Results.}
Figure~\ref{fig:dbps-overview} summarizes how our method augments a standard WebDataset pipeline with metadata-based filtering, bucket balancing, and rank-aware shard scheduling.
Table~\ref{tab:dbps} summarizes relative data utilization under distributed training on $W$ GPUs.
The offline bucket-balanced sampler with greedy shard assignment increases per-epoch throughput from $N$ to approximately $4.6N$, while utilization rises from roughly 20\% to roughly 76\%.
Adding SA further improves throughput to approximately $5.4N$, corresponding to about 18\% improvement over the greedy variant, with utilization approaching 90\%.
The remaining synchronization loss is modest and consistent with the discrete nature of clip-to-batch assignment.

\begin{table}[h]
  \centering
  \caption{Relative data utilization per epoch on $W$ GPUs.}
  \label{tab:dbps}
  \begin{tabular}{lccc}
    \toprule
    Method & Steps / epoch & Utilization & DL latency \\
    \midrule
    Baseline (clip shuffle)  & $N$ & $\approx 20\%$ & 0.05\,s/step \\
    Offline sampler + greedy         & $\approx 4.6N$ & $\approx 76\%$ & $<$0.001\,s/step \\
    Offline bucket sampler + SA \emph{(ours)}    & $\approx 5.4N$ & $\approx 90\%$ & $<$0.001\,s/step \\
    \bottomrule
  \end{tabular}
\end{table}

\paragraph{Outcome.}
In our setting, the offline bucket-balanced sampler provides a practical way to retain WebDataset's fast sequential reads while still supporting clip filtering and frame-resolution bucketing.
It substantially reduces synchronization loss and improves data-loading speed.

\section{Experiments}
\label{sec:experiments}

Our evaluation spans three levels: per-component design validation,
end-to-end benchmark comparison, and qualitative analysis. The first
category: attention-pattern analysis supporting the three-stage
design (Section~\ref{sec:architecture_stage},
Figure~\ref{fig:ddt-attn}), the text-attention dilution evidence
motivating Shared Cross-Attention
(Section~\ref{sec:shared-cross-attention}, Figure~\ref{fig:single_layer_attn_drop}),
the SkyReels-V4 stability comparison
(Section~\ref{sec:shared-cross-attention}, Figure~\ref{fig:scattn_vs_skyreels_1k}), and the
REPA teacher analysis (Section~\ref{sec:repa},
Figure~\ref{fig:jepa_vis}), is presented alongside the design
decisions it informs, following the convention that evidence is most
useful where the claim it supports is made. This section focuses on
the remaining two: quantitative evaluation on VBench
(Section~\ref{sec:exp:quant}) and qualitative results
(Section~\ref{sec:exp:qual}).

\subsection{Quantitative evaluation on VBench}
\label{sec:exp:quant}
Table~\ref{tab:vbench_main} reports VBench scores across all 16
dimensions. Unless otherwise noted in the table caption, scores are
reported from the public VBench leaderboard. Under the standard
open-source text-to-video setting, Motif-Video 2B achieves a Total
Score of 83.76\%, surpassing larger openly released models including
Wan2.1-T2V-14B (83.69\%), HunyuanVideo (83.24\%), and
Step-Video-T2V-30B (81.83\%). Wan2.2-T2V reports a higher total
score (84.23\%), but that entry uses prompt optimization, so we treat
it separately rather than as a like-for-like comparison.

The strongest gains for Motif-Video 2B are on the semantic side of the
benchmark. It leads open-source models with full per-dimension results
on Spatial Relationship (83.02\%), and ranks near the top on Object
Class (92.93\%), Multiple Objects (77.29\%), and overall Semantic
Score (80.44\%). This pattern is consistent with the paper's central
claim that the architecture prioritizes text grounding and
compositional control, especially for multi-object layouts and
spatially specified prompts. At the same time, the table shows clear
headroom on quality-related dimensions. Subject Consistency (95.38\%)
and Background Consistency (95.74\%) remain below the strongest Wan
models, and Temporal Flickering (98.16\%) trails the best scores in
the Wan2.1 family (up to 99.55\%). We therefore read the benchmark as
showing a specific trade-off rather than a uniform win: at 2B scale,
Motif-Video 2B is unusually strong on semantic alignment, while long
horizon temporal stability and appearance consistency remain the main
targets for further scaling and data improvement.


\begin{table*}[t]
\centering
\caption{%
  \textbf{VBench T2V evaluation across all 16 fine-grained dimensions (scores in \%).}
  \textbf{Bold} and \underline{underline} denote the best and second-best results
  among open-source models with full dimension scores, respectively.
  $^\dagger$~Closed-source; excluded from open-source rankings.
  $^p$~Evaluated with prompt optimisation
  (Qwen-rewritten prompts for Wan2.2; SAT-enhanced for CogVideoX1.5-5B).
  $^\alpha$~Updated HunyuanVideo API checkpoint (2025-05-22);
  open-source weights not released for this version.
  $^\star$~SANA-Video aggregate scores from our unified evaluation
  protocol~\cite{chen2025sana};
  individual dimension scores not publicly reported by the authors.
  $^\S$~Scores from the original paper~\cite{zheng2025open} under
  a T2I2V pipeline (FLUX anchor frame); individual dimension scores
  not reported. Direct comparison with standard T2V models
  should be interpreted with caution.
  All other scores sourced from the
  \textbf{VBench Leaderboard}~\cite{huang2024vbench}.
}
\label{tab:vbench_main}
\setlength{\tabcolsep}{2.4pt}
\renewcommand{\arraystretch}{1.15}
\resizebox{\linewidth}{!}{%
\begin{tabular}{l r c c c c c c c c c c c c c c c c c c c}
\toprule
\textbf{Model}
  & \textbf{Params}
  & \multicolumn{3}{c}{\textbf{Aggregate}}
  & \multicolumn{7}{c}{\textbf{Quality Dimensions}}
  & \multicolumn{9}{c}{\textbf{Semantic Dimensions}} \\
\cmidrule(lr){3-5}\cmidrule(lr){6-12}\cmidrule(lr){13-21}
 &
 & \rotatebox{60}{\small Total $\uparrow$}
 & \rotatebox{60}{\small Quality $\uparrow$}
 & \rotatebox{60}{\small Semantic $\uparrow$}
 & \rotatebox{60}{\small Subj.\ Cons.}
 & \rotatebox{60}{\small Bg.\ Cons.}
 & \rotatebox{60}{\small Temp.\ Flick.}
 & \rotatebox{60}{\small Motion Sm.}
 & \rotatebox{60}{\small Dyn.\ Degree}
 & \rotatebox{60}{\small Aesth.\ Qual.}
 & \rotatebox{60}{\small Img.\ Qual.}
 & \rotatebox{60}{\small Object Class}
 & \rotatebox{60}{\small Multiple Obj.}
 & \rotatebox{60}{\small Human Action}
 & \rotatebox{60}{\small Color}
 & \rotatebox{60}{\small Spatial Rel.}
 & \rotatebox{60}{\small Scene}
 & \rotatebox{60}{\small App.\ Style}
 & \rotatebox{60}{\small Temp.\ Style}
 & \rotatebox{60}{\small Ovr.\ Consist.} \\
\midrule
Veo~3~\cite{wiedemer2025video}$^\dagger$
 & -- & 85.06 & 85.70 & 82.49
 & 97.36 & 96.89 & 99.30 & 99.16 & 72.43 & 63.81 & 68.23
 & 93.89 & 82.20 & 99.40 & 82.48 & 84.26 & 57.43
 & 23.55 & 25.97 & 27.88 \\
Sora~\cite{brooks2024video}$^\dagger$
 & -- & 84.28 & 85.51 & 79.35
 & 96.23 & 96.35 & 98.87 & 98.74 & 79.91 & 63.46 & 68.28
 & 93.93 & 70.85 & 98.20 & 80.11 & 74.29 & 56.95
 & 24.76 & 25.01 & 26.26 \\
Luma$^\dagger$
 & -- & 83.61 & 83.47 & 84.17
 & 97.33 & 97.43 & 98.64 & 99.35 & 44.26 & 65.51 & 66.55
 & 94.95 & 82.63 & 96.40 & 92.33 & 83.67 & 58.98
 & 24.66 & 26.29 & 28.13 \\
HunyuanVideo$^{\dagger\alpha}$~\cite{kong2024hunyuanvideo}
 & -- & 83.43 & 85.07 & 76.88
 & 97.22 & 97.60 & 99.39 & 99.05 & 71.94 & 60.28 & 67.24
 & 83.48 & 66.71 & 94.40 & 89.79 & 72.13 & 54.46
 & 22.21 & 24.52 & 26.95 \\
MiniMax-Video-01$^\dagger$
 & -- & 83.41 & 84.85 & 77.65
 & 97.51 & 97.05 & 99.10 & 99.22 & 64.91 & 63.03 & 67.17
 & 87.83 & 76.04 & 92.40 & 90.36 & 75.50 & 50.68
 & 20.06 & 25.63 & 27.10 \\
Kling~1.6$^\dagger$
 & -- & 83.40 & 85.00 & 76.99
 & 97.40 & 96.84 & 98.64 & 99.13 & 62.22 & 64.81 & 69.70
 & 93.34 & 63.99 & 96.20 & 81.26 & 79.08 & 55.57
 & 20.75 & 24.51 & 26.04 \\
\midrule
Wan2.2-T2V$^p$~\cite{wan2025wan}
 & A14B & \textbf{84.23} & \underline{85.42} & 79.50
 & 97.29 & 97.39 & 99.22 & 98.16 & 61.02
 & \textbf{67.22} & \textbf{71.75}
 & \textbf{94.06} & \textbf{82.10} & 96.40 & 87.43 & 78.39 & \textbf{56.80}
 & 20.39 & 23.64 & 26.12 \\
SANA-Video$^\star$~\cite{chen2025sana}
 & 2B & 83.71 & 84.35 & \textbf{81.35}
 & -- & -- & -- & -- & -- & -- & --
 & -- & -- & -- & -- & -- & --
 & -- & -- & -- \\
Wan2.1-T2V~\cite{wan2025wan}
 & 14B & 83.69 & \textbf{85.59} & 76.11
 & 97.52 & \textbf{98.09} & \underline{99.46} & 98.30 & 65.46
 & \underline{66.07} & 69.43
 & 86.28 & 69.58 & 95.40 & 88.59 & 75.39 & 45.75
 & 22.64 & 23.19 & 25.91 \\
OpenSora~2.0$^\S$~\cite{zheng2025open}
 & 11B & 83.60 & 84.40 & 80.30
 & -- & -- & -- & -- & -- & -- & --
 & -- & -- & -- & -- & -- & --
 & -- & -- & -- \\
Wan2.1-T2V~\cite{wan2025wan}
 & 1.3B & 83.31 & 85.23 & 75.65
 & \underline{97.56} & \underline{97.93} & \textbf{99.55} & 98.52 & 65.19
 & 65.46 & 67.01
 & 88.81 & 74.83 & 94.00 & \underline{89.20} & 73.04 & 41.96
 & 21.81 & 23.13 & 25.50 \\
HunyuanVideo~\cite{kong2024hunyuanvideo}
 & 13B & 83.24 & 85.09 & 75.82
 & 97.37 & 97.76 & 99.44 & \underline{98.99} & \textbf{70.83}
 & 60.36 & 67.56
 & 86.10 & 68.55 & 94.40 & \textbf{91.60} & 68.68 & \underline{53.88}
 & 19.80 & 23.89 & 26.44 \\
CogVideoX1.5-5B$^p$~\cite{yang2024cogvideox}
 & 5B & 82.17 & 82.78 & 79.76
 & 96.87 & 97.35 & 98.88 & 98.31 & 50.93
 & 62.79 & 65.02
 & 87.47 & 69.65 & \underline{97.20} & 87.55 & \underline{80.25} & 52.91
 & \underline{24.89} & 25.19 & \underline{27.30} \\
CogVideoX-5B~\cite{yang2024cogvideox}
 & 5B & 81.91 & 83.05 & 77.33
 & 96.45 & 96.71 & 98.97 & 97.20 & \underline{69.51}
 & 61.88 & 63.33
 & 85.07 & 63.94 & \textbf{98.60} & 83.03 & 68.91 & 51.96
 & \textbf{24.98} & 25.42 & \textbf{27.65} \\
Step-Video-T2V~\cite{ma2025step}
 & 30B & 81.83 & 84.46 & 71.28
 & \textbf{98.05} & 97.67 & 99.40 & \textbf{99.08} & 53.06
 & 61.23 & \underline{70.63}
 & 80.56 & 50.55 & 94.00 & 88.25 & 71.47 & 24.38
 & 23.17 & \textbf{26.01} & 27.12 \\
LTX-Video~\cite{hacohen2024ltx}
 & 2B & 80.00 & 82.30 & 70.79
 & 96.56 & 97.20 & 99.34 & 98.96 & 54.35
 & 59.81 & 60.28
 & 83.45 & 45.43 & 92.80 & 81.45 & 65.43 & 51.07
 & 21.47 & 22.62 & 25.19 \\
\midrule
\textbf{Motif-Video 2B}~\textit{(Ours)}
 & \textbf{2B} & 83.76 & 84.59 & 80.44
 & 95.38 & 95.74 & 98.16 & 98.81 & 65.56
 & 65.95 & 70.50
 & 92.93 & 77.29 & 95.60 & 87.94 & 83.02 & 52.40
 & 23.03 & 25.24 & 26.78 \\
\bottomrule
\end{tabular}%
}
\end{table*}

\subsection{Effect of the Shared Cross-Attention}
\label{sec:exp-cross-attn}

Section~\ref{sec:shared-cross-attention} argues that Shared Cross-Attention injects text-derived
information that is both geometrically grounded in the backbone's existing key--value
manifold and directionally distinct from the self-attention output.
We verify this claim at inference time by probing all 16 single-stream encoder blocks
throughout the denoising trajectory (50 steps, $\sigma \in [1.00, 0.29]$,
1280$\times$736 at 121 frames, guidance scale~8).
At each block and step we record (i) the Frobenius norm of the cross-attention
contribution $\mathbf{W}_O^{\text{cross}}\,\text{Attn}(\mathbf{Q}, \mathbf{K}, \mathbf{V})$,
and (ii) its magnitude relative to the self-attention residual $\|\mathbf{h}_v\|$.

\begin{figure}[t]
    \centering
    \includegraphics[width=\linewidth]{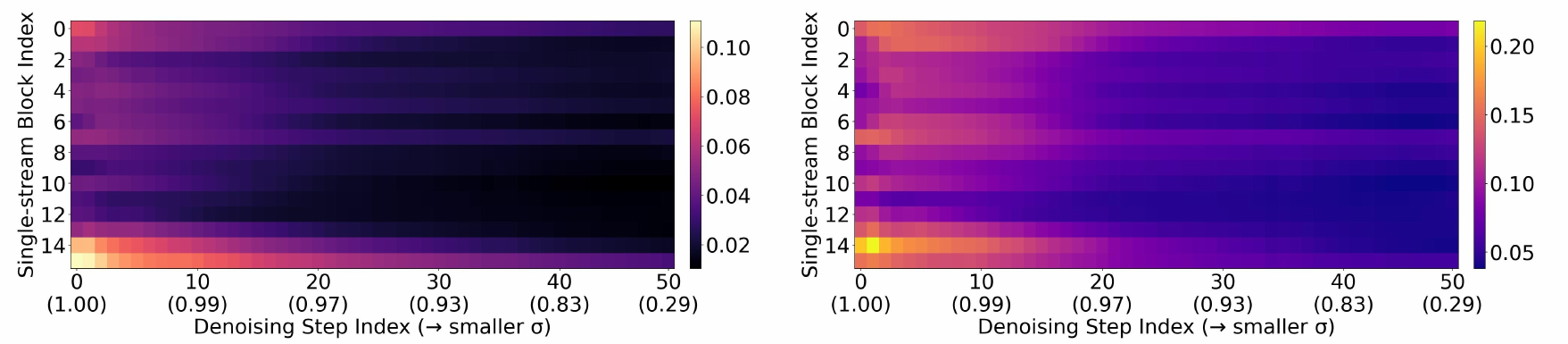}
    \caption{
        \textbf{Shared Cross-Attention contribution across single-stream encoder blocks
        and denoising steps} ($1280\times736$, 121 frames, 50 steps,
        $\sigma \in [1.00, 0.29]$).
        \emph{Left}: Frobenius norm of the cross-attention output
        $\mathbf{W}_O^{\text{cross}}\,\text{Attn}(\mathbf{Q},\mathbf{K},\mathbf{V})$
        per block (row) and step (column).
        \emph{Right}: ratio of the cross-attention residual norm to the self-attention
        output norm $\|\mathbf{h}_v\|$.
        No block falls below 5.2\%; the global mean is 7.6\% and the maximum 21.7\%.
        Activity peaks at block~0 and in the high-noise regime, stabilising after
        step~22 ($\sigma \approx 0.96$).
        The cross-attention residual is nearly orthogonal to the self-attention output
        (global cosine $\approx {-}0.008$), confirming that the module injects
        text-grounded information along directions absent from the self-attention pathway.
    }
    \label{fig:cross-attn-probe}
\end{figure}

\paragraph{Contribution magnitude.}
Figure~\ref{fig:cross-attn-probe} shows the per-block, per-step Frobenius norm heatmap.
The cross-attention signal is non-negligible across the full trajectory: globally it accounts
for \textbf{7.6\%} of the self-attention residual magnitude on average, rising to a maximum of
\textbf{21.7\%} (Figure~\ref{fig:cross-attn-probe}).
No block is dormant; the weakest contributes 5.2\%, confirming that all 16
cross-attention modules remain active participants rather than residual no-ops.
Block~0 is the most active (10.6\%), consistent with it receiving the least-processed
video hidden state and therefore drawing the most heavily on text for initial grounding;
blocks~14--15 follow (9.3--9.6\%), suggesting a final consolidation of text alignment just
before the DDT decoder receives the joint representation.

\paragraph{Directional orthogonality.}
Beyond magnitude, we measure the cosine similarity between the cross-attention
contribution and the self-attention output across all blocks and steps.
The global mean is $\cos(\mathbf{W}_O^{\text{cross}}\text{Attn},\,\mathbf{h}_v) \approx -0.008$:
the two residuals are nearly orthogonal.
This rules out the hypothesis that cross-attention acts as a signal amplifier or a
correction to existing self-attention features.
Instead, the module injects text information along directions that are almost entirely
absent from the self-attention output, a functional profile we term an \emph{information
injector} rather than a signal amplifier.
The orthogonality is consistent with the manifold argument in Section~\ref{sec:shared-cross-attention}:
by grounding $\mathbf{K}$ and $\mathbf{V}$ in the backbone's own text projections while
learning a free $\mathbf{Q}$ projection from the post-self-attention video state, the module
is positioned to ask a question the self-attention pathway was structurally unable to ask.

\paragraph{Denoising dynamics.}
The heatmaps reveal a step-wise pattern that the aggregate statistics obscure.
Cross-attention activity peaks in the high-noise regime ($\sigma \approx 1.0$), where global
semantic structure is being established, and stabilizes after step~22 ($\sigma \approx 0.96$),
when the convergence difference drops below 10\% of its peak value.
This trajectory mirrors the known dynamics of the flow-matching denoising process: early
steps are dominated by coarse semantic decisions to which text alignment is critical, while
later steps refine spatial detail in a regime where text influence is already baked into the
latent.
We note that the current experiment covers $\sigma \in [1.00, 0.29]$ only (shift~$= 20$);
the low-$\sigma$ tail ($\sigma < 0.25$) is not reached under this shift setting and
remains an open question for follow-up experiments with lower shift values.

Appendix~\ref{app:cross-attn-ablation} additionally compares generations with and without Shared Cross-Attention, and analyzes how the module changes the observed contribution patterns. Consistent with the analysis above, these contribution patterns are reflected in the generated videos themselves: removing Shared Cross-Attention leads to visibly weaker prompt alignment and less coherent scene realization.

\subsection{Qualitative results}
\label{sec:exp:qual}

\begin{figure*}[t]
    \centering
    \includegraphics[width=\textwidth]{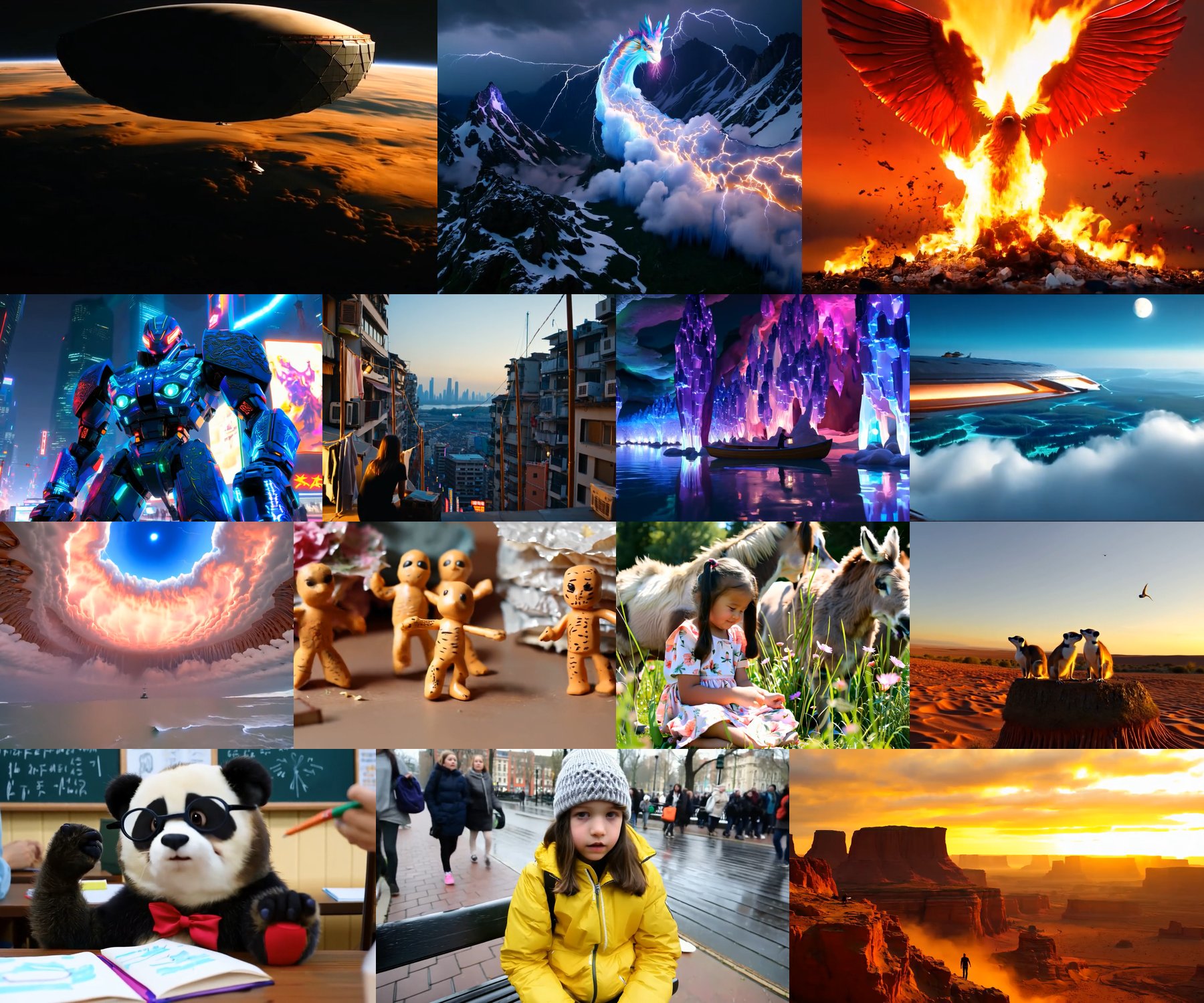}
    \caption{\textbf{Selected single-frame samples from Motif-Video 2B across a
    range of subjects and visual styles.} Each tile is a frame drawn from an
    independently generated text-to-video clip. The grid is intended to
    convey the breadth of domains the model handles, including photographic scenes,
    stylized and fantastical content, close-up subjects, and wide
    landscapes, rather than to claim uniform quality across all prompts.}
    \label{fig:mosaic}
\end{figure*}

We present qualitative samples from Motif-Video 2B for text-to-video
generation. Figure~\ref{fig:main_banner} shows multi-frame strips from a set of
prompts that stress temporal behavior: camera motion, subject articulation,
and scene dynamics. The strips are chosen to make temporal coherence
inspectable at a glance: neighboring frames should read as a continuous
clip rather than as independently sampled images. Figure~\ref{fig:mosaic}
complements this with a wider grid of single frames drawn from diverse
prompts, illustrating the range of subjects, styles, and compositions that
the model covers under a single set of weights.

Both figures are curated. We select them to communicate what Motif-Video 2B
does well, not to characterize its average behavior; the VBench breakdown
in Table~\ref{tab:vbench_main} serves that purpose. Motif-Video 2B also exhibits
characteristic failure modes, most visibly in fine-grained human anatomy
and in long-horizon temporal stability; we discuss these directly in
Sec~\ref{sec:dis:limitations}. 

To complement these curated examples,
Appendix Figure~\ref{fig:additional-video-human} presents additional
un-curated generations involving human subjects, providing a broader view
of the model's typical video outputs on prompts that are especially
sensitive to anatomical fidelity and motion consistency.

\begin{figure*}[t]
    \centering
    \includegraphics[width=\linewidth]{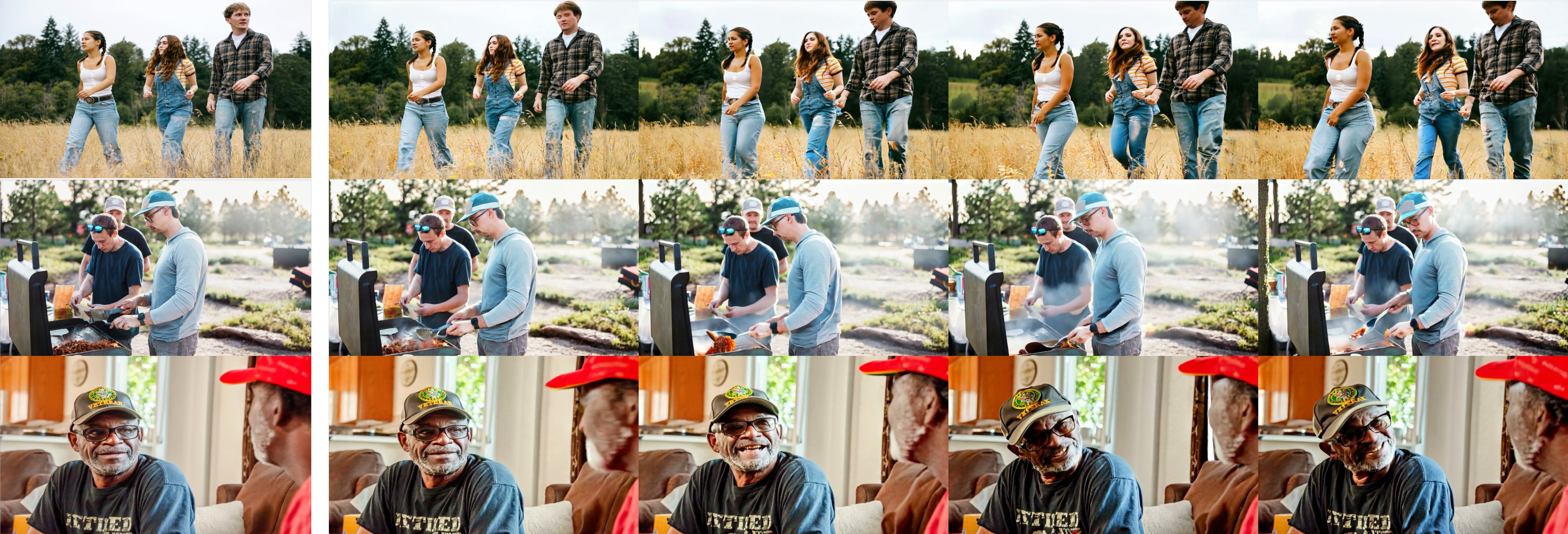}
    \caption{\textbf{Image-to-video generation results.} The leftmost panel is the input image, and the model preserves its original appearance while generating temporally coherent video content from it.}
    \label{fig:i2v-main}
\end{figure*}

We also verify that Motif-Video 2B supports image-to-video generation.
Figure~\ref{fig:i2v-main} shows that the model can animate a given
input image while preserving its original appearance and scene structure.
The leftmost panel is the input image, and the generated frames confirm
that the image-to-video capability is learned without losing fidelity to
the source image. Appendix Figure~\ref{fig:additional-i2v-0}
presents additional image-to-video results.

\subsection{Human evaluation}
\label{sec:exp:human}

\begin{table}[t]
\centering
\caption{Human evaluation results. Pairwise preferences are converted to
ELO ratings. \emph{Total} aggregates all
pairwise judgments across both axes; \emph{prompt-following} measures
whether the generated video matches the input text;
\emph{video-fidelity} measures visual coherence and plausibility,
independent of the prompt. Models are sorted by Total ELO.}
\label{tab:human-eval}
\small
\begin{tabular}{lcccc}
\toprule
Model & Params & Total $\uparrow$ & Prompt-following $\uparrow$ & Video-fidelity $\uparrow$ \\
\midrule
Wan2.1            & 14B        & 1114.8 & 1054.4 & 1101.6 \\
HunyuanVideo-1.5  & 6.5B   & 1083.0 & 1045.2 & 1043.8 \\
Wan2.2            & 5B         & 1062.4 & 1039.1 & 1049.9 \\
LTX-Video~2       & 14B & 1041.5 & 1005.0 & 1043.7 \\
\textbf{Motif-Video (ours)} & \textbf{2B} & \textbf{1026.2} & \textbf{1015.1} & \textbf{1047.3} \\
Wan2.1            & 1.3B       & 1015.6 & 1010.1 & 1006.5 \\
SANA-Video        & 2B   &  915.3 &  943.5 &  948.5 \\
CogVideoX         & 5B         &  851.5 &  916.5 &  872.3 \\
\bottomrule
\end{tabular}
\end{table}

Automatic benchmarks such as VBench aggregate many dimensions into a
single score, but they correlate only loosely with what a human viewer
actually perceives as a good video.
To complement the automatic evaluation in Sec~\ref{sec:exp:quant},
we run a blind pairwise study that targets the two qualities we care
about most: whether the generated video matches the prompt, and whether
it looks coherent and visually plausible.

We evaluate on a set of $40$ prompts generated by an LLM. To avoid
biasing the prompts toward any particular model's strengths, we
condition the LLM on a public prompting
guide\footnote{\url{https://docs.ltx.video/api-documentation/prompting-guide}}
rather than on examples from our own training distribution.

We compare Motif-Video~2B against six contemporaneous open-source
video generators spanning a wide range of parameter counts and
training-data scales: SANA-Video~\citep{chen2025sana},
LTX-Video~2~\citep{hacohen2026ltx}, Wan2.1-14B and Wan2.1-1.3B~\citep{wan2025wan},
Wan2.2-5B, and CogVideoX-5B~\citep{yang2024cogvideox}.
For every baseline we use the recommended default inference
configuration published on its Hugging Face model card (sampler,
guidance scale, step count, resolution, frame count).

For each prompt, every pair of models produces one video, and an
annotator is shown the two clips side by side with the prompt
displayed above. Model identities and left/right order are randomized
and hidden. Annotators answer two independent questions:
\emph{prompt-following} (``which video better matches the text
description?'') and \emph{video-fidelity} (``which video looks more
coherent and visually plausible, ignoring the prompt?''). We
deliberately separate the two axes because, as we will see, models
can rank very differently on each.

\paragraph{Results.}

\begin{figure}[t]
    \centering
    \includegraphics[width=\linewidth]{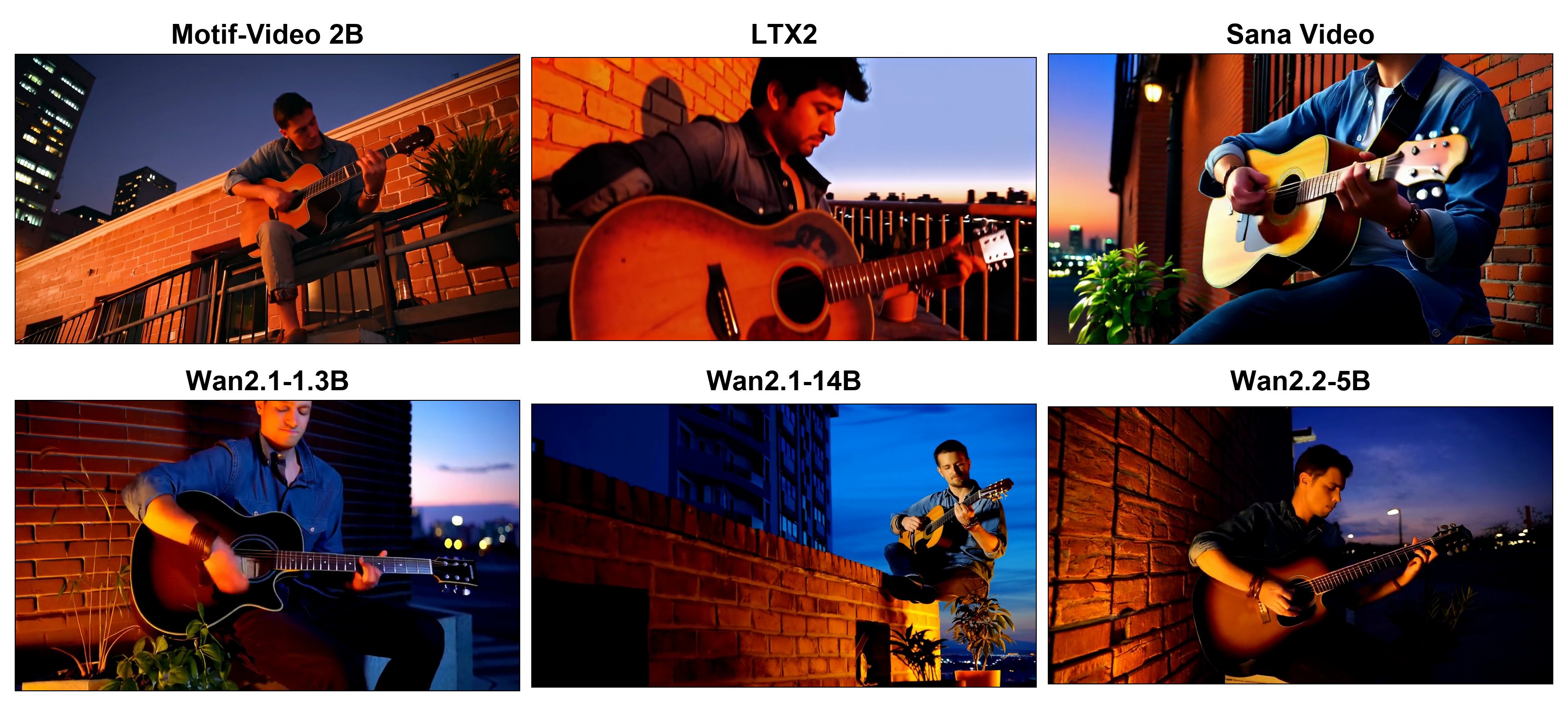}
    \caption{\textbf{Example of generated results from the arena.} Prompt: \emph{"A guitarist sits on a fire escape playing at twilight, fingers moving in relaxed patterns along the neck of a scratched acoustic guitar. Shot on a 40mm lens with a slow crane-up from the street below, the brick wall beside him glows deep orange as the last sun hits it and the sky above shifts toward indigo. He wears a loose denim shirt rolled to the elbows, a leather bracelet knocking softly against the guitar body. A potted plant beside him sways in the warm updraft. The camera rises past him to reveal the skyline beginning to glitter with early window lights."} Note that even high-fidelity models sometimes fail on certain examples, but we have observed such models, for example Wan2.1 14B, consistently show better results, unlike the example shown here.}
    \label{fig:arena-comparison}
\end{figure}

Table~\ref{tab:human-eval} reports ELO ratings on both axes. Two
observations stand out. First, the picture differs sharply from the VBench ranking in
Sec~\ref{sec:exp:quant}. Wan2.1-14B is preferred over
Motif-Video~2B by a clear margin on both prompt-following and
video-fidelity, despite Motif-Video~2B holding the high VBench
Total Score among open-source models. Inspecting the failure cases,
we find that Motif-Video~2B and other models near its scale most
often lose due to \emph{semantic} failures---missing or swapped
subjects, ignored attributes, prompt--video mismatches---rather than
to low-level visual artifacts. We discuss the implications of this
gap, and what it suggests about the limitations of uniformly weighted
benchmark aggregates, in Sec~\ref{sec:discussion}.

Second, within the comparable scale regime, Motif-Video~2B is preferred
over both SANA-Video (similar parameter count) and Wan2.1-1.3B
(similar parameter count, substantially larger training corpus) on
both axes. We read this as evidence that the architectural and
training-recipe choices described in
Sections~\ref{sec:model_architecture}--\ref{sec:training_strategy} translate into
perceptible quality gains at fixed scale, rather than merely improving
benchmark scores.

\paragraph{Caveats.}
A $40$-prompt study with the annotator pool described above is
sufficient to surface the qualitative picture reported here, but it
is not large enough to support fine-grained claims about small
ELO differences. In particular, we do not interpret the ranking
\emph{among} models within overlapping confidence intervals as
meaningful, and we do not claim that Motif-Video~2B is uniformly
better than any baseline it outranks---only that, under matched
default-inference conditions, human raters tend to prefer it. A
larger, more controlled study---with a broader prompt distribution,
more annotators per pair, and per-dimension breakdowns of failure
modes---is left to future work.

\section{Discussion}
\label{sec:discussion}

\subsection{Interpretation of Results}

The results in Section~\ref{sec:experiments} admit more than one
reading. Before turning to the boundaries of what our recipe can do,
we discuss what we believe the results say, and what they do not
say, about the design choices in this report.

\paragraph{The gap between VBench and perceptual quality.}
Motif-Video 2B reaches the highest Total Score among open-source
models we evaluate (83.76\%, Table~\ref{tab:vbench_main}), but in our
internal side-by-side comparisons against Wan2.1-T2V-14B we observe a
consistent perceptual gap in favor of the larger model, despite
trailing it by only 0.07 points on the aggregate metric. We take this
seriously and report it explicitly.

Two factors contribute to the discrepancy. First, VBench weights its
sixteen dimensions uniformly in the aggregate score, whereas human
perceptual preference is disproportionately sensitive to temporal
stability: a viewer forgives a missing object more readily than a
flickering one, but VBench penalizes both equally. Second, VBench's
semantic dimensions can award credit for near-correct outputs,
for instance, a generated human whose anatomy is subtly distorted but
who performs the prompted action in the correct spatial configuration
will score well on Human Action, Spatial Relationship, and Object
Class, even though a human viewer would immediately flag the
anatomical artifact. Motif-Video 2B's strength on these semantic
dimensions is genuine, but the scores do not fully distinguish between
``semantically correct'' and ``perceptually convincing''.

A fairer parameter-class comparison is Wan2.1-T2V-1.3B (83.31\%).
Against this baseline Motif-Video 2B leads by 0.45 points on Total
Score and by 4.79 points on Semantic Score (80.44\% vs.\ 75.65\%),
while the two models trade wins on quality dimensions: Wan2.1-1.3B
holds an edge on Subject Consistency (97.56\% vs.\ 95.38\%) and
Temporal Flickering (99.55\% vs.\ 98.16\%), while Motif-Video 2B
leads on Aesthetic Quality (65.95\% vs.\ 65.46\%) and Imaging Quality
(70.50\% vs.\ 67.01\%).
Even this comparison is not fully controlled: Wan2.1 reports training
on billions of images and videos~\cite{wan2025wan}, roughly two orders of
magnitude more data than the fewer than 10M clips used by
Motif-Video 2B.
In internal side-by-side evaluation at the $\sim$2B scale, the two
models are substantially closer in perceived quality than either is
to Wan2.1-14B.

We therefore interpret our VBench result not as a claim that
Motif-Video 2B matches Wan2.1-14B in perceived quality, but as
evidence that a 2B model trained under our recipe can match a 14B
model on the compositional and semantic axes of generation while
remaining capacity-limited on the temporal-stability axes, and that
within its own parameter class, the recipe yields a clear advantage on
semantic understanding without sacrificing quality parity.

\paragraph{Data as the ceiling on an efficient design.}
The architectural and training choices in this report are designed to
maximize what a fixed data budget can deliver, and the semantic
results in Table~\ref{tab:vbench_main} suggest they succeed at this:
a 2B model trained on fewer than 10M clips matches or exceeds 14B
models trained on one to two orders of magnitude more
data~\cite{wan2025wan,kong2024hunyuanvideo} on compositional dimensions.
But design efficiency does not remove the need for data; it lowers
the threshold at which data becomes the binding constraint.
We believe Motif-Video 2B has reached that threshold.
The long-tail domain gaps and dynamic-motion degradation described
below are, in our assessment, symptoms of data coverage rather than
architectural limitations; unlike the image domain, where
hundreds of millions of captioned pairs are publicly available,
high-quality video data with diverse motion and temporal coherence
remains scarce.
Scaling the training corpus in quantity, motion diversity, and domain
breadth is the most natural next step, and one that the current
architecture is positioned to absorb.

\paragraph{Scaling outlook.}
The three-stage backbone and Shared Cross-Attention are
parameter-agnostic designs: nothing in their formulation ties them to
2B. We expect the role-separation philosophy to remain useful at
larger scales, but the \emph{optimal} allocation across stages may
shift. In particular, the DDT decoder currently uses 8 layers,
roughly 22\% of the total depth, and our analysis suggests that
temporal coherence is concentrated there.  A natural scaling
experiment is to hold the encoder fixed and grow the decoder, testing
whether the consistency gap closes before the semantic advantage
erodes.  We view this, together with the data scaling direction
above, as the most immediate paths for a future iteration.

\subsection{Limitations}
\label{sec:dis:limitations}

\begin{figure*}[t]
    \centering
    \includegraphics[width=\textwidth]{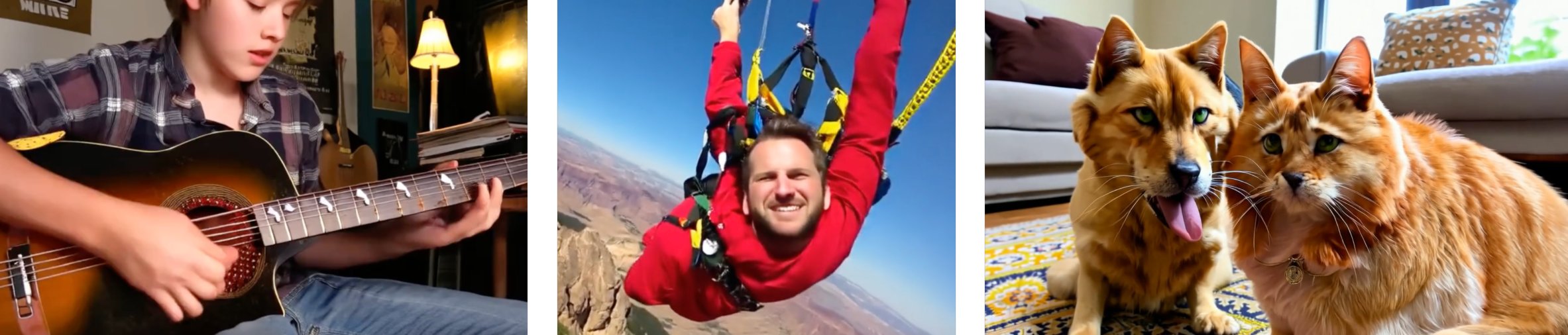}
    \caption{\textbf{Micro-scale semantic distortion.} Three characteristic
    failures at the sub-object level: distorted hand anatomy on a close-up
    instrument subject (left), broken body structure under a high-motion
    skydiving prompt (middle), and attribute leakage between co-present
    animals in a multi-subject scene (right). The generations may remain
    category-correct (guitar, skydiver, cat and dog), leading VBench's semantic
    dimensions award credit, but a human viewer flags the artifact on first
    inspection.}
    \label{fig:failure_semantic}
\end{figure*}

\begin{figure*}[t]
    \centering
    \includegraphics[width=\textwidth]{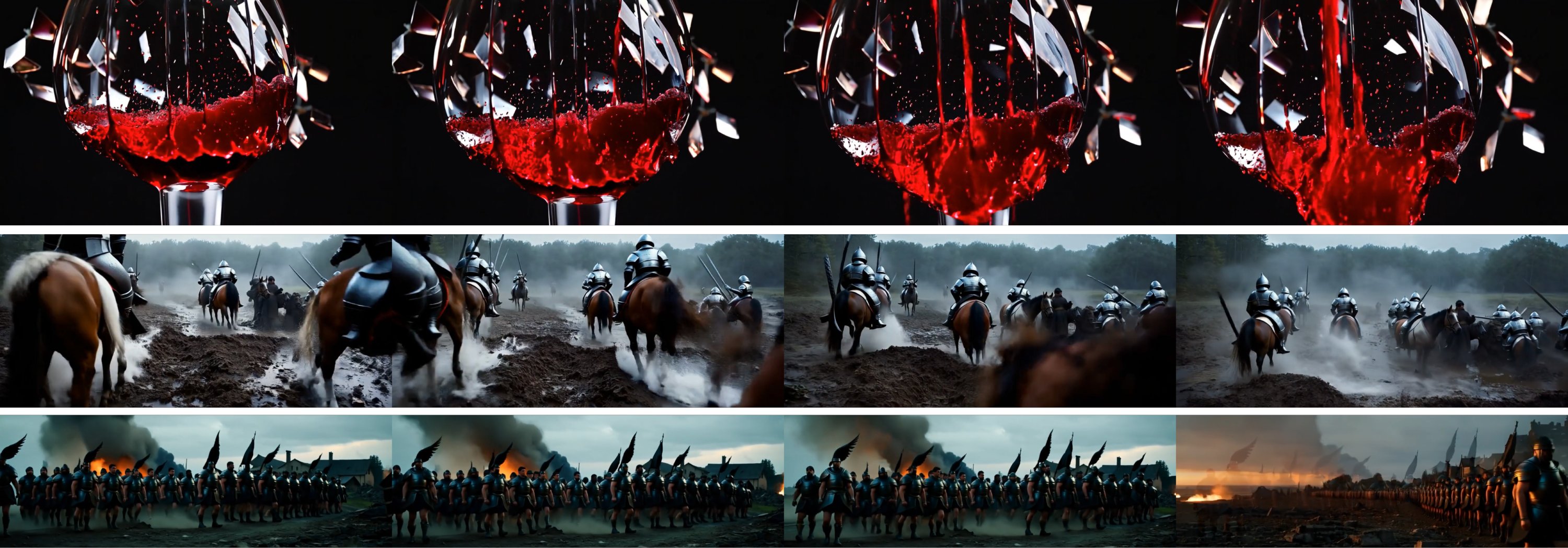}
    \caption{\textbf{Temporal failure modes.} Top: physically implausible
    liquid dynamics in a wine-splash prompt: the motion is locally smooth
    but violates gravity and surface tension. Middle: loss of temporal
    coherence under high scene complexity in a cavalry-charge prompt, where
    subject identities blur across frames and multi-agent spatial
    relationships fail to persist. Bottom: unintended mid-clip scene
    transition, where the prompted setting drifts into an unrelated
    composition partway through the sequence.}
    \label{fig:failure_temporal}
\end{figure*}

We report limitations not as caveats but as the boundary conditions
under which the design decisions in this report should be interpreted.
Several of them point directly to follow-up work; others are
properties of the 2B operating regime that scaling is the most likely
remedy for.

\paragraph{Failure modes.}
The VBench aggregate in Table~\ref{tab:vbench_main} captures
compositional and semantic correctness, but it is largely insensitive
to two classes of perceptual failure that a human viewer notices
immediately. We document both directly in
Figures~\ref{fig:failure_semantic} and~\ref{fig:failure_temporal} and
discuss what each suggests for a future iteration.

\textit{Micro-scale semantic distortion
(Figure~\ref{fig:failure_semantic}).}
Motif-Video 2B occasionally produces sub-object-level artifacts that
leave the category label intact but break perceptual plausibility:
distorted hands on close-up human subjects, degraded body structure
under high-displacement motion, and attribute leakage between
co-present subjects of similar size and color. These failures are
consistent with the VBench-to-perception gap discussed in
Section~\ref{sec:discussion}: the prompted objects are present in
the correct spatial configuration, so the aggregate score is largely
unaffected, but the artifacts are immediately visible on direct
inspection. We attribute them primarily to data coverage rather than
to the backbone design. Fine-grained anatomical fidelity and robust
multi-subject disambiguation scale with the quantity and diversity of
training clips covering the relevant subject, and a sub-10M corpus is
thin in exactly the regions where these failures concentrate:
close-up human extremities, high-displacement body motion, and
multi-animal scenes with visually similar subjects.

\textit{Temporal failures (Figure~\ref{fig:failure_temporal}).}
We observe three distinct temporal failure modes that a static-frame
metric cannot surface. The first is physical implausibility:
generated liquids, cloth, and rigid-body collisions can evolve
smoothly frame-to-frame while violating gravity, surface tension, or
momentum conservation. The second is coherence loss under high scene
complexity: in dense multi-agent prompts such as the cavalry charge
in Figure~\ref{fig:failure_temporal} (middle), subject identities
blur across frames and the spatial relationships established in the
opening frames fail to persist. The third is unintended scene
transitions, in which the model drifts mid-clip from the prompted
setting into an unrelated composition. These failures do not share a
single cause. Physical plausibility is fundamentally a data
question: without sufficient exposure to physics-rich clips, no
amount of temporal capacity will recover the correct dynamics from
the flow-matching objective alone. Complex-scene coherence and
within-clip consistency, in contrast, are more plausibly
capacity-bound, and are the failures most likely to benefit from the
decoder-side scaling direction noted in
Section~\ref{sec:discussion}.

\paragraph{Recipe components are evaluated jointly, not in isolation.}
We do not present per-component ablations for Shared Cross-Attention,
the DDT decoder, REPA phasing, or TREAD routing. The empirical
evidence we provide is the attention-pattern analysis
(Figures~\ref{fig:ddt-attn},~\ref{fig:single_layer_attn_drop}) and the
SkyReels-V4 vs.\ Shared Cross-Attention comparison
(Figure~\ref{fig:scattn_vs_skyreels_1k}), together with the end-to-end
VBench result. A cleaner attribution of contribution-per-component
would require ablation training runs at the same scale, which we did
not have the compute budget to perform. Readers should interpret our
results as evidence that the \emph{composed} recipe works at 2B, not
as a claim about the marginal contribution of any single component.

\paragraph{Open questions in the training recipe.}
Two specific questions about the recipe remain unresolved. First, we
disable REPA at the 360p transition based on the phase-constrained
alignment argument of~\cite{wang2025repa}, but we have not tested whether
a holistic, early-stopped variant in the spirit of HASTE would extend
the useful lifetime of the alignment signal in our setting. Second,
our V-JEPA 2.0 teacher provides spatially fragmented dense features
(Figure~\ref{fig:jepa_vis}), and we expect that a teacher with denser
spatial structure (e.g., V-JEPA 2.1~\cite{mur2026v}) would change the
trade-off of when REPA should be turned off. Both questions are
natural extensions rather than corrections.

\section{Conclusion}
\label{sec:conclusion}

This report asks whether competitive text-to-video generation
requires massive scale, and presents evidence that it does not,
provided that model design explicitly separates the objectives that
scaling would otherwise leave entangled.
Motif-Video 2B reaches 83.76\% on VBench with 2B parameters, fewer
than 10M training clips, and under 100,000 H200 GPU hours, matching
or exceeding models 7$\times$ its size on compositional and semantic
dimensions.
Three design choices drive this result: Shared Cross-Attention
stabilizes text conditioning under the token imbalance inherent to
long video sequences; the three-stage backbone assigns modality
fusion, joint representation, and detail reconstruction to dedicated
components; and the DDT decoder, applied to video for the first time,
develops inter-frame attention structure that the encoder alone does
not exhibit.
On the training side, the combination of TREAD token routing and
phase-constrained REPA with a V-JEPA teacher, which to our knowledge was
first composed for video diffusion, delivers a micro-budget recipe
in which a 27\% per-step FLOP reduction coexists with structured
early-phase learning, and an offline bucket-balanced sampler recovers
90\% data utilization from a baseline of 20\%.
Temporal stability and data coverage remain the primary constraints;
the former is localized in the decoder by the same role-separation
design, and the latter defines the most natural scaling axis for a
future iteration that the current architecture is built to absorb.

\bibliographystyle{plainnat}
\bibliography{references}

\clearpage

\appendix

\section{Additional results}

\begin{figure}[t]
    \centering
    \includegraphics[width=\linewidth]{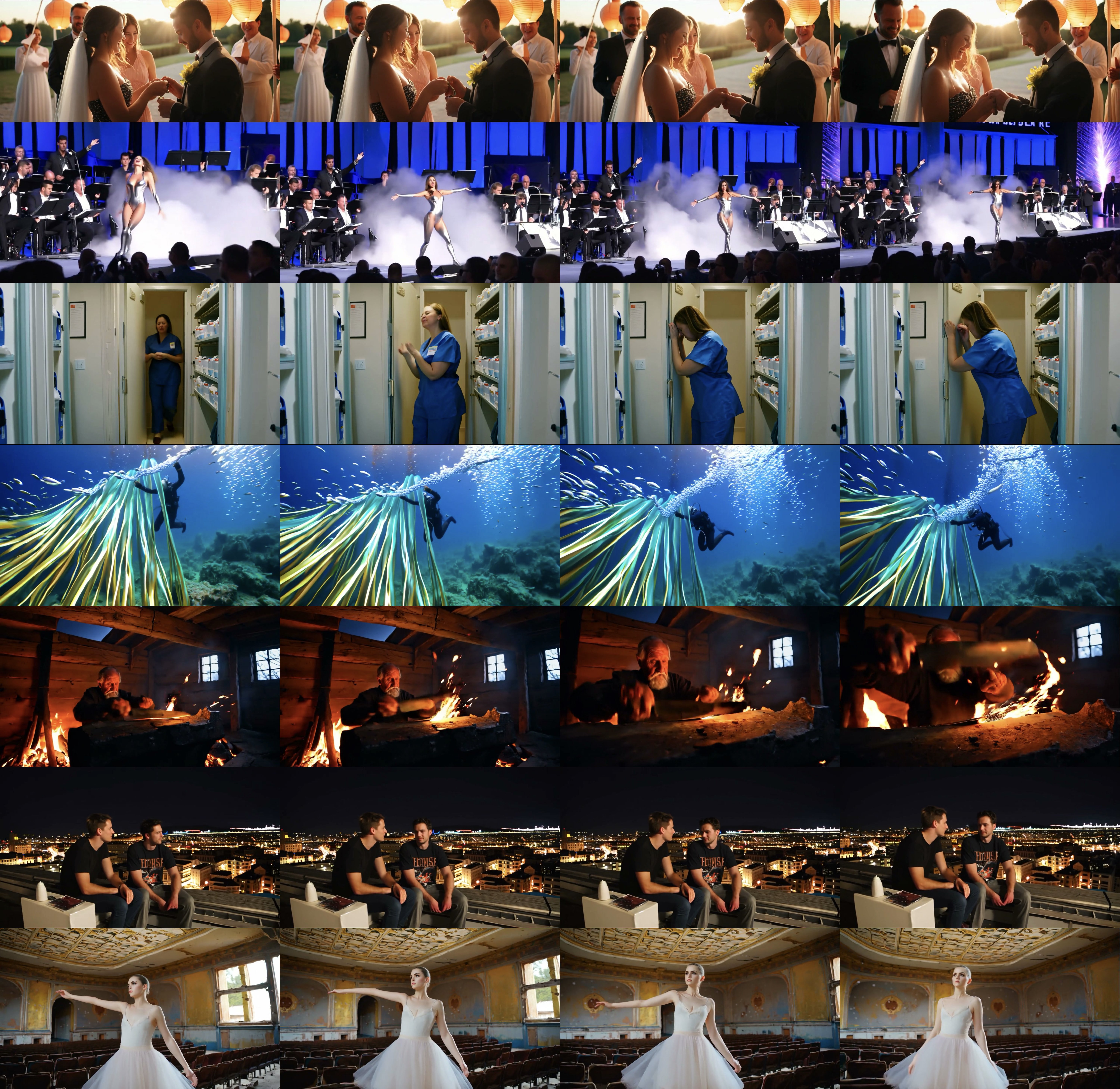}
    \caption{\textbf{Additional qualitative human-centered generations.} Representative frames from videos involving human subjects, included as supplementary qualitative results.}
    \label{fig:additional-video-human}
\end{figure}

\begin{figure}[t]
    \centering
    \includegraphics[width=\linewidth]{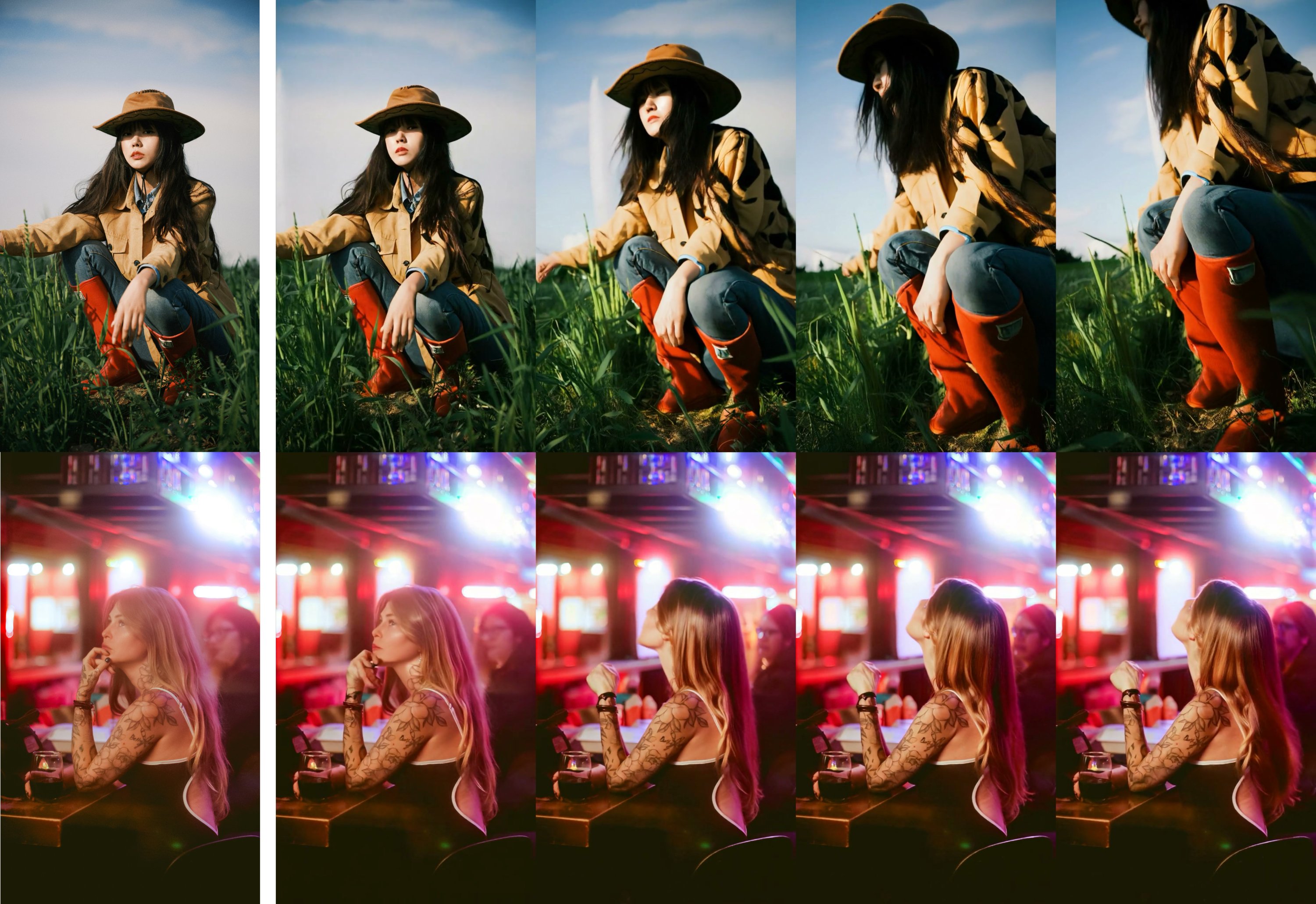}
    \caption{\textbf{Additional image-to-video results.} The leftmost panel is the input image, and the remaining panels show representative generated video frames.}
    \label{fig:additional-i2v-0}
\end{figure}

This section presents additional qualitative results for both text-to-video and image-to-video generation in Figures~\ref{fig:additional-video-human} and~\ref{fig:additional-i2v-0}.

\section{Sampling Configuration}
\label{app:sampling}

We describe the sampling configuration used to produce the VBench scores
reported in Table~\ref{tab:vbench_main}. All samples are generated at
$1280 \times 736$ spatial resolution, $121$ frames, and $24$ fps.

\paragraph{Sampler.}
Following Waver~\citep{zhang2025waver}, we use the Video APG
sampler~\citep{sadat2024eliminating} with momentum $=0.0$, $\eta=0.0$, and $r=27$. The
classifier-free guidance scale is set to $8$.

\paragraph{Timestep schedule.}
We adopt the linear-quadratic timestep schedule proposed by Meta
Movie~Gen~\citep{polyak2024movie}, with the linear-to-quadratic transition point set
to $t=250$.

\paragraph{Negative prompt.}
Following Wan~\citep{wan2025wan}, we apply a fixed negative prompt at every
sampling call. The full string used is:

\begin{quote}\small\ttfamily
The video has text and graphic overlays burned into the frame, including 
watermarks, logos, subtitles, timestamps, broadcast graphics, UI elements, 
and stray letters scattered in corners and center. The subject stays nearly 
frozen in a rigid pose with minimal gesture or expression change, and the 
little motion present looks jerky, mechanical, and discontinuous between 
frames. The framing feels flat, rigid, and depthless. The lighting is dull 
and monotone, with crushed shadows in dark regions and blown-out highlights 
in bright regions at the same time. The background fades out, shifts into 
an unrelated scene, and loses continuity without any smooth transition. 
The subject's identity drifts across frames, with deformation, flickering 
detail, ghosting, smearing, and duplication in the face, body proportions, 
clothing, and accessories. Colors are flat, desaturated, and tonally 
compressed, and the foreground blends into the background without clear 
separation. Brightness, exposure, and color balance shift unevenly between 
consecutive frames.
\end{quote}

\section{Training Configuration}
\label{app:training-config}

\paragraph{Optimizer.}
We use AdamW with $\beta_1=0.9$, $\beta_2=0.99$, $\epsilon=10^{-8}$, and
weight decay $0.0$. Gradients are globally clipped to a maximum norm of
$1.0$.

\paragraph{Learning rate.}
Rather than committing to a fixed schedule, we adjust the learning rate
adaptively across stages based on a combination of qualitative inspection,
VBench scores, and the current training resolution. Whenever the training
configuration changes in a way that perturbs the loss landscape, such as introducing
Shared Cross-Attention, increasing the spatial or temporal resolution, or
otherwise altering the model or data distribution, we apply a short linear
warmup before resuming training at the target learning rate.

\paragraph{Timestep sampling.}
At training time we first draw $u \sim \mathcal{U}[0,1]$ and then transform
$u$ into a training timestep $t$ using one of two distributions, depending
on the current resolution stage. For resolutions below $360$p, we use
the logit-normal density of \citet{esser2024scaling},
\begin{equation}
    \pi_{\mathrm{ln}}(t; m, s)
    = \frac{1}{s\sqrt{2\pi}} \, \frac{1}{t(1-t)}
      \exp\!\left( -\frac{(\mathrm{logit}(t) - m)^2}{2 s^2} \right),
\end{equation}
with $(m, s) = (0, 1)$. From $360$p onward, we switch to the cosine
mode-sampling map of \citet{esser2024scaling},
\begin{equation}
    f_{\mathrm{mode}}(u; s)
    = 1 - u - s \cdot \!\left( \cos^2\!\left( \tfrac{\pi}{2} u \right) - 1 + u \right),
\end{equation}
with $s = 1.29$.
The early-stage logit-normal places more density near the high-noise
region, which we found stabilizes training when the model is still
learning coarse structure at low resolution; the mode-sampling
distribution then redistributes density toward intermediate timesteps
once the model is operating at higher resolutions where mid-noise
denoising dominates perceptual quality.

\paragraph{Resolution-dependent timestep shifting.}
On top of the sampled $t$, we apply the standard rectified-flow timestep
shift $t \mapsto \tfrac{\sigma t}{1 + (\sigma - 1) t}$, with the shift
factor $\sigma$ increased adaptively as resolution grows, up to a maximum
of $\sigma = 7.0$ at our highest training resolution. Larger shifts bias
sampling toward higher-noise timesteps, which we found necessary to
preserve global structure as the token count per sample increases.

\paragraph{Classifier-free guidance dropout.}
To enable classifier-free guidance at inference, we independently drop
the text and image conditions with probability $10\%$ each during
training. For dropped samples, the unconditional input is constructed by
re-encoding the empty string \texttt{""} through the T5Gemma2 text
encoder, rather than by substituting a zero tensor. We found that
zero-tensor unconditioning produces a malformed unconditional score
estimate and degrades CFG sample quality, whereas encoding the empty
string yields an unconditional distribution that is consistent with the
encoder's output manifold.

\section{Implementation Details for FSDP2 Wrapping Order}
\label{app:fsdp2_impl}

\paragraph{Activation checkpointing, compilation, and FSDP wrapping order.}
We apply activation checkpointing, \texttt{torch.compile}, and FSDP2
\texttt{fully\_shard} in that order. Per-block checkpoint wrapping must
precede compilation so that compile regions align with the
activation-checkpointed units, and both must precede FSDP so that each
checkpointed, compiled block is sharded as an independent FSDP unit with
its own parameter all-gather and gradient reduce-scatter. Wrapping is
applied at three granularities: each individual transformer block, the
enclosing transformer module, and the root model.

Accelerate's built-in FSDP2 path does not support this ordering for our
model. Its \texttt{fsdp2\_apply\_ac} locates activation-checkpointing
targets through the FSDP auto-wrap policy applied to \emph{parent}
modules, an indirection that does not reliably hit the individual blocks
inside our \texttt{transformer\_blocks} and
\texttt{single\_transformer\_blocks} \texttt{ModuleList}s. We therefore
patch two entry points. First, \texttt{fsdp2\_apply\_ac} is replaced
with a version that directly iterates both \texttt{ModuleList}s, applies
\texttt{checkpoint\_wrapper} to each child block, and re-registers the
wrapped block into its parent via \texttt{register\_module} so that it
replaces the original in place. Second,
\texttt{Accelerator.\_prepare\_fsdp2} is patched to enforce the
activation-checkpointing $\rightarrow$ compile $\rightarrow$ FSDP
sequence above; the stock implementation interleaves these steps in a
way that breaks once activation checkpointing is applied at block
granularity. To have FSDP shard the checkpoint-wrapped blocks as
independent units, we include \texttt{CheckpointWrapper} in
\texttt{transformer\_cls\_names\_to\_wrap} alongside
\texttt{MotifVideoTransformer3DModel}, producing the three-level
wrapping hierarchy described above.

\section{Cross-Attention Ablation Details}
\label{app:cross-attn-ablation}

\begin{figure*}[htbp]
    \centering
    \begin{subfigure}{\linewidth}
        \includegraphics[width=\linewidth]{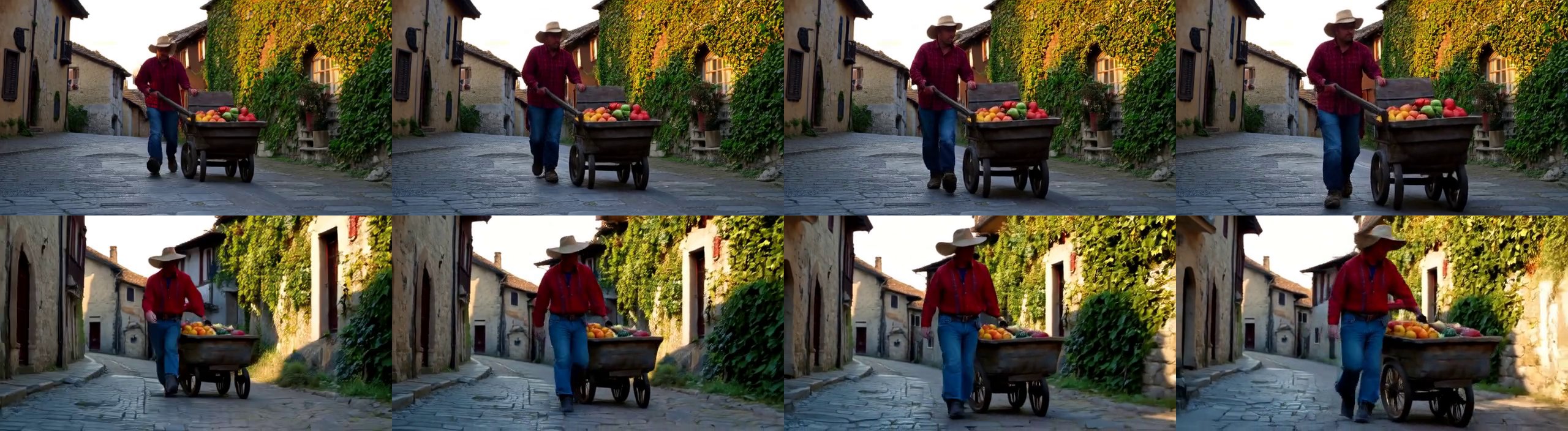}
        A determined individual, dressed in a red flannel shirt, blue jeans, and sturdy boots, \textcolor{red}{pushes a weathered wooden cart} along a narrow, cobblestone street. The scene is set in a quaint, old-world village with charming stone buildings and ivy-covered walls. The cart, filled with an assortment of colorful fruits and vegetables, creaks slightly as it moves. The person's face, partially obscured by a wide-brimmed hat, shows a mix of focus and determination. As they push the cart, the early morning sun casts long shadows, adding a golden hue to the scene, while birds chirp softly in the background, enhancing the serene atmosphere.
        \label{fig:sub1}
    \end{subfigure}
    
    \vspace{1.5em} 
    
    \begin{subfigure}{\linewidth}
        \includegraphics[width=\linewidth]{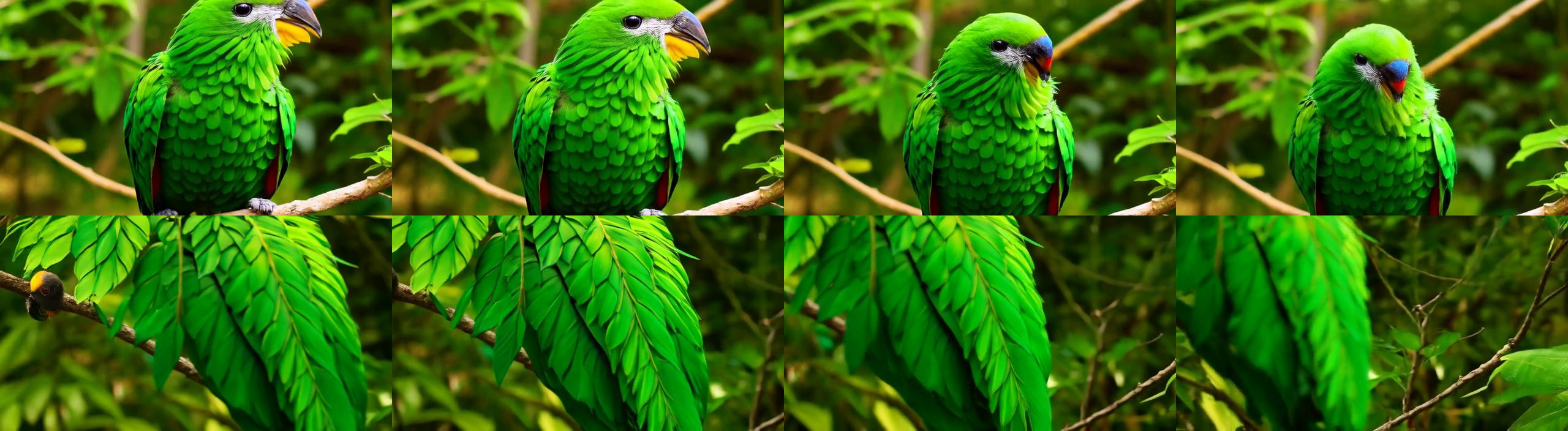}
        A vibrant \textcolor{red}{green parrot with iridescent feathers} perches on a delicate branch in a lush rainforest, its eyes gleaming with curiosity. The camera zooms in to capture the intricate details of its plumage, each feather shimmering in shades of emerald and lime. The bird tilts its head, revealing a striking yellow patch on its cheek, and lets out a melodious chirp that echoes through the dense foliage. As it flutters its wings, the sunlight filters through the canopy, casting a dappled glow on its vivid colors. The scene transitions to the parrot taking flight, its wings spreading wide, gliding gracefully through the verdant landscape, embodying the essence of freedom and natural beauty.
        \label{fig:sub2}
    \end{subfigure}
    
    \vspace{1.5em} 
    
    \begin{subfigure}{\linewidth}
        \includegraphics[width=\linewidth]{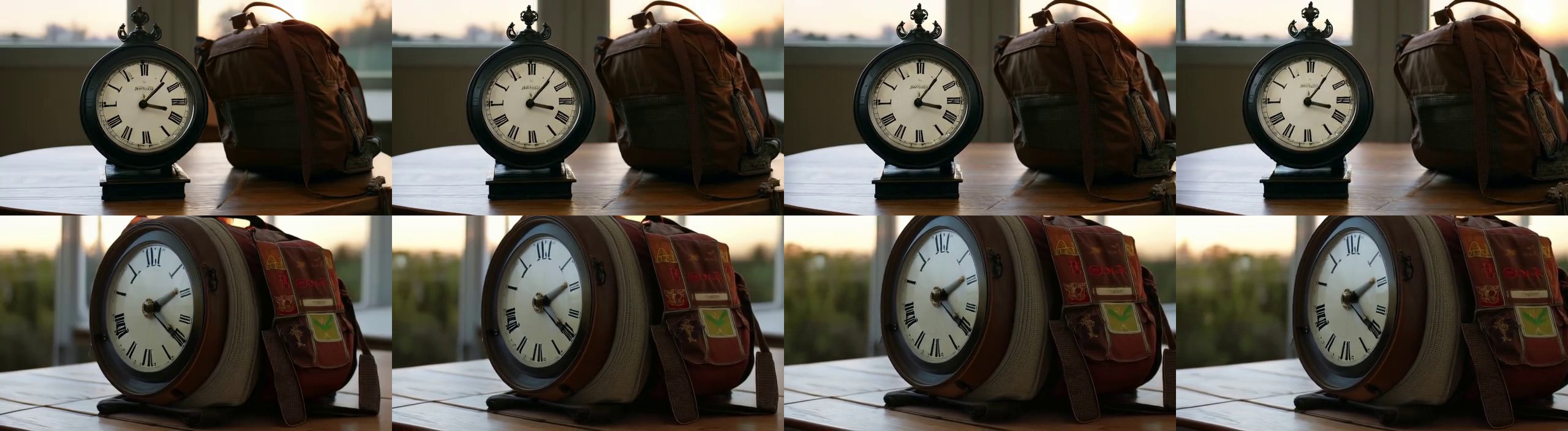}
        \textcolor{red}{A vintage clock with ornate hands} and Roman numerals sits on a rustic wooden table, its ticking sound filling the air. Beside it, a \textcolor{red}{well-worn leather backpack}, adorned with travel patches and a slightly frayed strap, leans against the table. The clock's face reflects the soft morning light streaming through a nearby window, casting gentle shadows. The backpack, partially open, reveals a glimpse of a map and a journal, hinting at adventures past and future. The scene evokes a sense of nostalgia and wanderlust, with the clock symbolizing the passage of time and the backpack representing the journey ahead.
        \label{fig:sub3}
    \end{subfigure}
    
    \caption{\textbf{Qualitative effect of Shared Cross-Attention.} For each prompt, the top row shows generation with Shared Cross-Attention enabled; the bottom row shows the same prompt and seed with cross-attention disabled on all 16 single-stream encoder blocks (360p, 50 steps, 121 frames).}
    \label{fig:cross-attn-ablation}
\end{figure*}


The qualitative ablation in Figure~\ref{fig:cross-attn-ablation} is performed by disabling \texttt{enable\_text\_cross\_attention} at runtime in all 16 single-stream encoder blocks and running the full 50-step denoising trajectory from the same initial noise. We use the checkpoint from Stage~9 (Section~\ref{sec:training_strategy}) and generate samples at $640 \times 360$ resolution, with 121 frames and a fixed seed. The dual-stream and DDT decoder blocks are unchanged, as they do not include Shared Cross-Attention.

Without cross-attention, text conditioning degrades in qualitatively distinct ways depending on the prompt's compositional demands. For ``A person is pushing cart,'' the enabled model correctly depicts a person \emph{pushing} the cart forward; the disabled model reverses the action, showing the person \emph{pulling} the cart, while the scene layout is preserved but the verb semantics are lost. For ``A green bird,'' the failure is more severe: the disabled model renders green foliage but \emph{no bird at all}, capturing the adjective while entirely dropping the noun. For ``A clock and backpack,'' the disabled model \emph{merges the two objects} into a single hybrid form rather than placing them as distinct entities, collapsing the compositional structure of the prompt. These three failure modes, verb confusion, noun loss, and object merging, are precisely the symptoms the softmax dilution analysis in Section~\ref{sec:shared-cross-attention} predicts: without a dedicated text-conditioning pathway, fine-grained semantic distinctions are overwhelmed by the video-dominated attention budget.

\section*{Contributions}

All authors are alphabetically sorted by last name.

\paragraph{Core contributors.} Wai Ting Cheung\footnotemark[1], Minsu Ha\footnotemark[1], Beomgyu Kim\footnotemark[1], Taewhan Kim\footnotemark[1], Haesol Lee\footnotemark[1], Dongpin Oh\footnotemark[1], Jeesoo Lee\footnotemark[2], Taehyun Kim\footnotemark[2], Minjae Kim\footnotemark[3]
\footnotetext[1]{model design, evaluation, training, data processing}
\footnotetext[2]{system optimization (kernel, parallelization)}
\footnotetext[3]{infrastructure}

\paragraph{Technical and management leadership.} Sungmin Lee, Junghwan Lim

\paragraph{Contributors.} Hyeyeon Cho, Dahye Choi, Jaeheui Her, Jaeyeon Huh, Hanbin Jung, Changjin Kang, Dongseok Kim, Jangwoong Kim, Youngrok Kim, Hyukjin Kweon, Hongjoo Lee, Jeongdoo Lee, Junhyeok Lee, Eunhwan Park, Yeongjae Park, Bokki Ryu, Dongjoo Weon

\paragraph{Acknowledgement} We gratefully acknowledge SkyPilot~\citep{yang2023skypilot} for helping us manage large-scale training infrastructure efficiently, and NVIDIA NeMo Curator~\citep{jennings2024nemo} for providing a practical and scalable data-curation toolkit that substantially streamlined our preprocessing pipeline.

\section*{Use of Large Language Models}

In preparing this report, we used large language model (LLM) based
assistants for English language editing, including grammar correction,
rephrasing for clarity, and improving the readability of passages
originally drafted by the authors. LLMs were not used to generate
research ideas, design experiments, produce or analyze results, write
code, or draft technical content beyond such surface-level editing.
All scientific claims, experimental findings, and figures in this
report were produced and verified by the authors, who take full
responsibility for the contents of the manuscript.

\end{document}